\documentclass[11pt]{videococo}
\usepackage{enumitem}
\usepackage{float}
\usepackage{needspace}
\newfontfamily\arxivkeepbytesansmedium[Path=./]{ByteSans-Medium.ttf}
\newfontfamily\arxivkeepbytesansbold[Path=./]{ByteSans-Bold.ttf}

\newcommand{\guigenbrand}{%
  \textcolor{gradstart!100!gradend}{G}%
  \textcolor{gradstart!91!gradend}{U}%
  \textcolor{gradstart!82!gradend}{I}%
  \textcolor{gradstart!73!gradend}{-}%
  \textcolor{gradstart!64!gradend}{G}%
  \textcolor{gradstart!55!gradend}{e}%
  \textcolor{gradstart!46!gradend}{n}%
  \textcolor{gradstart!37!gradend}{B}%
  \textcolor{gradstart!28!gradend}{e}%
  \textcolor{gradstart!19!gradend}{n}%
  \textcolor{gradstart!10!gradend}{c}%
  \textcolor{gradend}{h}%
}

\fancypagestyle{firststyle}{%
  \fancyhf{}
  \fancyhead[L]{%
    \makebox[22mm][c]{\includegraphics[height=\instlogoheight]{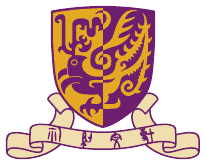}}%
    \hspace{5mm}%
    \makebox[22mm][c]{\includegraphics[height=\instlogoheight]{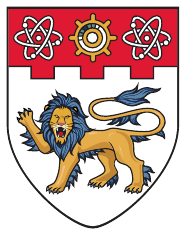}}%
  }
  \fancyhead[R]{%
    \makebox[22mm][c]{\includegraphics[height=\instlogoheight]{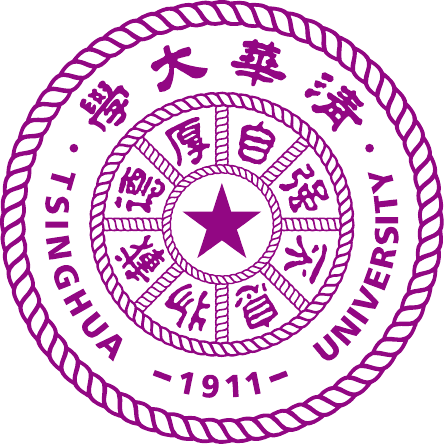}}%
    \hspace{5mm}%
    \makebox[22mm][c]{\includegraphics[height=\instlogoheight]{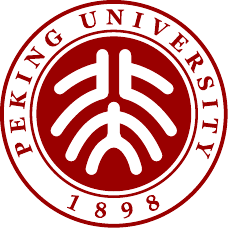}}%
  }

}
\hypersetup{
  pdftitle={GUI-GenBench: Evaluating Image Generation Models as Interactive GUI Environments},
  pdfauthor={GUI-GenBench authors},
  pdfsubject={Local arXiv layout draft}
}

\setlist{itemsep=0.25em,topsep=0.35em,parsep=0pt,partopsep=0pt}
\correspondence{\email{mickyhimself4@gmail.com}}
\checkdata[Project Page]{\url{https://github.com/stepfun-ai/GEBench}}

\definecolor{firstcolor}{HTML}{F49767}  % 最深
\definecolor{secondcolor}{HTML}{F9BA86} % 中等
\definecolor{thirdcolor}{HTML}{F5D5AC}  % 最浅
\definecolor{goldcolor}{HTML}{F49767}
\definecolor{silvercolor}{HTML}{F5D5AC}

\usepackage{mathtools}
\usepackage{amsthm}

\theoremstyle{plain}

\theoremstyle{definition}

\theoremstyle{remark}

\title{\guigenbrand: Evaluating Image Generation Models as Interactive GUI Environments}
\author[1]{Haodong Li}
\author[2]{Jingwei Wu}
\author[2,\dagger]{Quan Sun}
\author[2]{Guopeng Li}
\author[3]{Juanxi Tian}
\author[1]{\\Ziyu Guo}
\author[1]{Donghao Zhou}
\author[4]{Tianfei Ren}
\author[5]{Huanyu Zhang}
\author[6]{Yanlin Lai}
\author[7]{\\Ruichuan An}
\author[2]{Hongbo Peng}
\author[2]{Yuhong Dai}
\author[2]{Jia Wang}
\author[2]{Ziyang Meng}
\author[2]{\\Zheng Ge}
\author[2]{Xiangyu Zhang}
\author[2]{Daxin Jiang}
\author[1,*]{Pheng-Ann Heng}

\renewcommand{\affiliationlist}{{\affiliationfont\sffamily
  $^{1}$CUHK\quad
  $^{2}$StepFun\quad
  $^{3}$NTU\quad
  $^{4}$USTC\quad
  $^{5}$CASIA\quad
  $^{6}$THU\quad
  $^{7}$PKU
}}
\contribution[\dagger]{Project leader}
\contribution[*]{Corresponding author.}

\abstract{%

An image generation model used as a graphical user interface (GUI) environment must produce not only a plausible screen but also the correct response to an action. Existing visual-generation benchmarks do not directly assess this combination of functional correctness and temporal consistency in GUIs.
We introduce \textbf{GUI-GenBench}, a benchmark of 700 curated samples across five task suites: single-step transitions, multi-step planning, fictional-app generation, real-app trajectories, and coordinate-based grounding. We also introduce \textbf{GUI-Score}, a VLM-based metric that assesses Goal Achievement, Interaction Logic, Content Consistency, UI Plausibility, and Visual Quality.
Evaluating 12 image generation models on Chinese and English subsets, we find that strong single-step performance does not consistently extend to coherent multi-step trajectories or accurate grounding. The highest aggregate GUI-Scores are 69.62 on the Chinese subset and 63.16 on the English subset. Qualitative analysis further identifies failures in text rendering, icon interpretation, and coordinate-conditioned transitions. Together, these results distinguish visual plausibility from functional reliability and identify limitations that must be addressed before generated GUIs can serve as dependable interaction environments.

}

\begin{document}
\raggedbottom
\maketitle
\clearpage

% Color definitions for tables
\definecolor{colorfirst}{RGB}{252,141,89}
\definecolor{colorsecond}{RGB}{253,187,132}
\definecolor{colorthird}{RGB}{253,212,158}
\definecolor{colorfourth}{RGB}{254,232,200}
\definecolor{colorfifth}{RGB}{255,247,236}
\definecolor{myred}{RGB}{242,128,128}
\definecolor{mygreen}{RGB}{112,180,143}
\definecolor{myblue}{RGB}{210,225,255}
\definecolor{citypink}{RGB}{227,108,194}
\definecolor{cityblue}{RGB}{128,159,225}

\newcommand{\ph}[1]{\textcolor{black}{#1}}
\newcommand{\rankfirst}[0]{\cellcolor{colorfirst}}
\newcommand{\ranksecond}[0]{\cellcolor{colorsecond}}
\newcommand{\rankthird}[0]{\cellcolor{colorthird}}
\newcommand{\rankfourth}[0]{\cellcolor{colorfourth}}
\newcommand{\rankfifth}[0]{\cellcolor{colorfifth}}
\DeclareRobustCommand{\legendsquare}[1]{%
  \textcolor{#1}{\rule{2ex}{2ex}}%
}
\newcommand{\cmark}{\textcolor{mygreen}{\checkmark}}%
\newcommand{\xmark}{\textcolor{myred}{\times}}%

\section{Introduction}
\label{sec:introduction}

Image generation models~\citep{hurst2024gpt, comanici2025gemini, team2023gemini, seedream2025seedream, flux-2-2025, wan2025} can synthesize a new graphical user interface (GUI) state from a current screen and a user instruction. This capability motivates their use as \textbf{generative GUI environments}: models that produce visual responses to actions, with potential applications in agent training and evaluation~\citep{zhang2025large, yan2025gui, xie2025mirage, garg2025controllable, luo2025vimo, wei2023boosting}. Conventional interaction environments depend on implemented dynamics or executable software~\citep{cobbe2020leveraging, xie2024osworld, bonatti2024windows, zhang2025large}. Generating screens offers a complementary way to construct diverse interface examples and trajectories~\citep{zhao2021guigan, liu2025ui}, but raises a basic question: do the generated states faithfully reflect the requested interaction?

\begin{figure}[t]
    \centering
    \includegraphics[width=\linewidth]{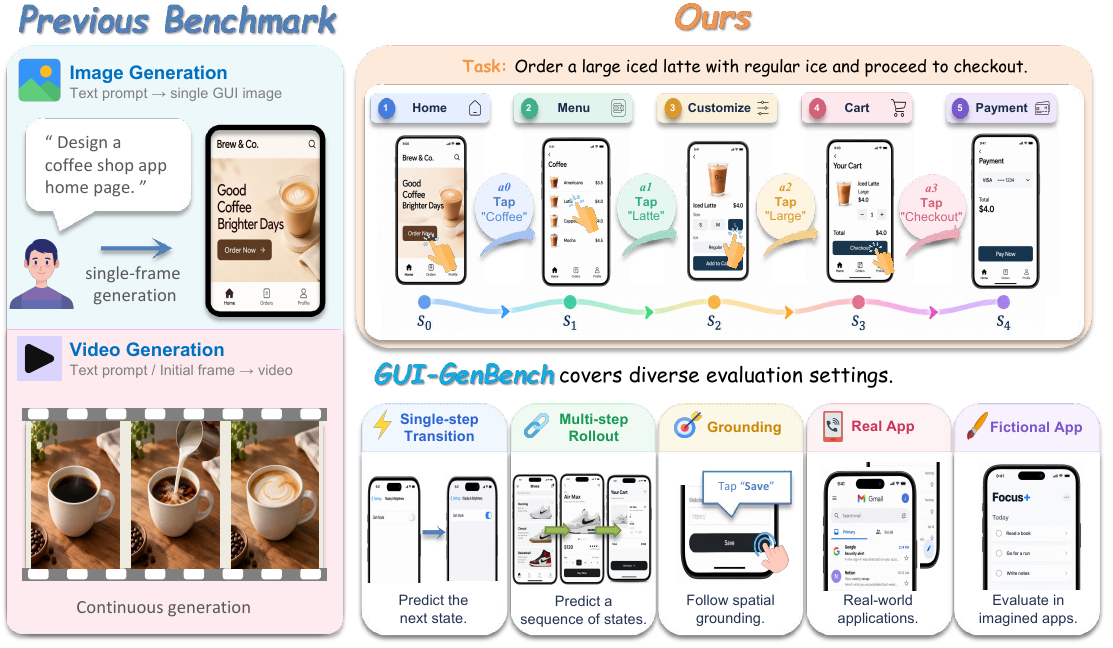}
    % \vspace{-3mm}
    \caption{\textbf{GUI-GenBench overview.} Image- and video-generation benchmarks assess visual content and temporal behavior. GUI-GenBench focuses on whether generated GUI states respond correctly to actions and remain coherent across a trajectory. Its five task suites cover single-step transitions, multi-step planning, grounding, real-app trajectories, and fictional-app generation.}
    \label{fig:teaser_fig}
    % \vspace{-3mm}
\end{figure}

Existing work on visual generation and evaluation examines fidelity, semantic alignment, composition, and temporal consistency~\citep{ghosh2023geneval, hu2024ella, niu2025wise, zhao2025envisioning, huang2023t2i, huang2024vbench, sun2025t2v, zhuang2025vistorybench}. These properties alone do not establish the functional correctness of GUI transitions~\citep{xie2025gui, yan2025gui}. As Figure~\ref{fig:teaser_fig} illustrates, a GUI environment must model discrete, action-triggered changes~\citep{zhang2025large}: a single click can replace an entire screen, while a small text or control-state error can change the meaning of the response. Correct transitions require coordinate grounding~\citep{zhao2021guigan, cheng2024seeclick}, icon interpretation~\citep{liu2025ui, xie2025gui}, and readable text~\citep{chen2024textdiffuser}, together with logical continuity across changing layouts~\citep{li2025mobileworldbench}. Evaluation must therefore test both what a screen looks like and whether it represents the intended state.

We introduce \textbf{GUI-GenBench} to evaluate image generation models along these functional and visual requirements. Its 700 curated samples span five task suites, from local state transitions to multi-step trajectories, fictional scenarios, and coordinate-based interactions. We pair the benchmark with \textbf{GUI-Score}, a rubric-based metric evaluated by vision-language models (VLMs)~\citep{google2025gemini3, hurst2024gpt, bai2025qwen3vltechnicalreport}. Five scoring dimensions separate goal achievement and interaction logic from content consistency, UI plausibility, and visual quality. This distinction makes it possible to identify outputs that look convincing but fail to realize the requested action.

Our evaluation covers 12 image generation models~\citep{openai2025gptimage, team2023gemini, seedream2025seedream, seedream2025seed, flux-2-2025, wan2025, wu2025qwen, deng2025emerging, li2025uniworld, team2025longcat}. Leading models score above 80 on single-step transitions, yet most models obtain lower scores on multi-step planning. Grounding remains difficult even when the generated screen is visually plausible. Qualitative examples reveal errors in Chinese text rendering, icon interpretation, and action-target localization. These findings motivate evaluating functional reliability separately from visual fidelity, rather than treating realistic appearance as evidence of correct interaction.

Our primary contributions are as follows:
\begin{enumerate}
\setlength{\itemsep}{0.25em}
\setlength{\topsep}{0.35em}
\setlength{\parsep}{0pt}
\item We introduce \textbf{GUI-GenBench}, with 700 samples across five task suites for evaluating generated GUI states and trajectories.
\item We introduce \textbf{GUI-Score}, a five-dimensional rubric-based metric that separates functional correctness, consistency, and visual fidelity.
\item We evaluate 12 models on Chinese and English subsets, identifying limitations in multi-step coherence and grounding, with qualitative diagnoses of text, icon, and coordinate-related errors.
\end{enumerate}

\section{Related Work}
\label{sec:related}

\subsection{Automated GUI Generation}

Work on automated interface construction spans layout design, screenshot-to-code translation, and generative prototyping~\citep{chen2018ui, sandhaus2011employing, yang2016automatic, li2020layoutgan, zhao2021guigan, mozaffari2022ganspiration, sobolevsky2023guilget, kolthoff2025guide, kolthoff2024zero}. Early approaches use explicit layout principles or reconstruct interfaces from mockups~\citep{sandhaus2011employing, huang2016automaticly}; learned approaches map screenshots to implementation structure~\citep{chen2018ui} or generate layouts and prototypes~\citep{sobolevsky2023guilget, kolthoff2025guide}. Generative layout and design methods broaden the space of candidate interfaces~\citep{li2020layoutgan, zhao2021guigan, mozaffari2022ganspiration}, while spatial reasoning and visual representation remain relevant challenges for multimodal models~\citep{zhang2025scaling,zhang2025latent}. GUI-GenBench focuses on a different evaluation target: whether a generated interface correctly changes in response to an instruction.

\subsection{Image Generation and Editing Models}

Recent multimodal systems support image synthesis and reference-based editing~\citep{hurst2024gpt, comanici2025gemini, team2023gemini, seedream2025seedream, wan2025, deng2025emerging, liu2025step1x, li2025uniworld, team2025nextstep, team2025longcat}. Diffusion and autoregressive approaches have advanced visual generation~\citep{ho2020denoising, chen2020generative, flux-2-2025, fan2024fluid, han2025infinity, ramesh2022hierarchical}, alongside progress in visual understanding~\citep{lin2025perceiveanythingrecognizeexplain}. Reference-conditioned methods combine visual context with textual instructions to control generated content~\citep{team2023gemini, seedream2025seedream, wan2025, team2025longcat, an2025unictokens}, including preservation of appearance and structure during editing~\citep{deng2025emerging, liu2025step1x}. These capabilities make GUI state generation feasible, but do not by themselves establish that the outputs obey interface logic.

\subsection{Visual and Sequential Generation Benchmarks}

Image-generation evaluation considers distributional fidelity, text--image alignment, and compositional correctness~\citep{heusel2017gans, ghosh2023geneval, huang2023t2i, clip_metrics2021}. Related work examines knowledge-informed generation, semantic alignment, and instruction-driven editing~\citep{niu2025wise, hu2024ella, zhao2025envisioning, huang2023t2i, zhang2026vibe-benchmark, liu2025step1x}, while sequential benchmarks study story consistency and visual reasoning~\citep{zhuang2025vistorybench, guo2025video}. GUI trajectories introduce a distinct requirement: actions must induce semantically correct changes even when the entire layout is replaced~\citep{zhang2025large, yan2025gui}. Text-rendering errors can also change the functional meaning of a screen~\citep{chen2024textdiffuser}. GUI-GenBench therefore evaluates the semantic and structural correctness of transitions, rather than visual continuity alone.

\subsection{Relation to Generative GUI Environment Work}

A related line of work constructs synthetic GUI data for agent training. UI-Oceanus~\citep{wu2026uioceanus} uses forward-dynamics prediction as a continual-pretraining objective, generating future GUI states from current observations and actions. GUI-ReWalk~\citep{lin2025guirewalk} combines stochastic exploration with intent-aware reasoning to construct interaction trajectories for downstream agents. GUI-GenBench complements these efforts by evaluating the generator rather than downstream agent success. Its five scoring dimensions provide a diagnostic of generated states that can inform synthetic-data pipelines. We do not measure whether training on these outputs improves agent performance.

\begin{figure*}[t]
    \centering
    \includegraphics[width=0.90\textwidth]{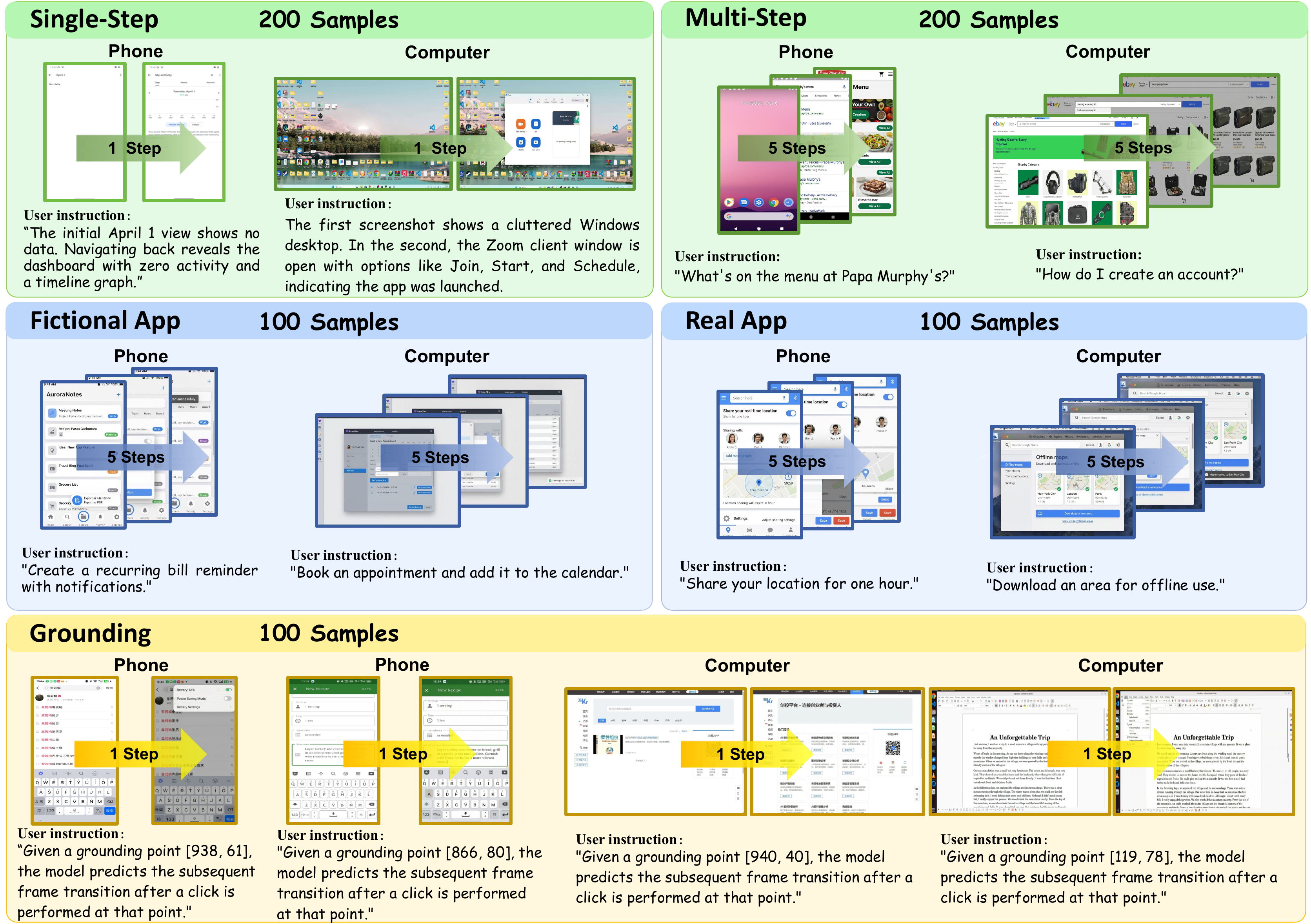}
    % \vspace{-5mm}
    \caption{\textbf{Examples from the five GUI-GenBench task suites.} Models generate GUI states or trajectories from task instructions and, where specified, a reference screen. The fictional-app task requires generation without a reference image. Evaluation considers both the requested interaction and the resulting visual content.}
    % \vspace{-2mm}
    \label{fig:benchmark_case}
\end{figure*}

\section{GUI-GenBench}
\label{sec:benchmark}

A GUI environment maps user or agent actions to interface states according to application logic~\citep{zhang2025large}. GUI-GenBench evaluates an image-based approximation of this process: a model receives an instruction and, when required by the task, a reference screen, then generates the resulting screen or trajectory. Throughout the paper, ``GUI state'' refers to the rendered image, not an executable application state. The benchmark tests whether these images depict valid responses; it does not verify execution in the underlying application.

\subsection{Benchmark Design and Task Suites}
\label{sec:tasks}

GUI-GenBench comprises 700 samples curated with consistency and fidelity checks. The five task suites in Figure~\ref{fig:benchmark_case} probe different requirements of GUI generation:

\begin{enumerate}

    \setlength{\itemsep}{0.25em}
    \setlength{\topsep}{0.35em}
    \setlength{\parsep}{0pt}

    \item \textbf{Single-step Visual Transition (single-step).} Given an initial GUI state and a detailed action, the model generates the subsequent state. This task evaluates fine-grained instruction following.

    \item \textbf{Multi-step Planning (multi-step).} Given an initial GUI state and a high-level objective such as ``Order a coffee,'' the model generates a five-step trajectory. This task evaluates goal-directed state progression, temporal coherence, and consistency of interface structure.

    \item \textbf{Zero-shot Virtual GUI (fictional-app).} The model generates a fictional interface from detailed instructions without a reference image. This task probes generation beyond familiar application layouts; it does not establish that the requested layout is absent from training data.

    \item \textbf{Rare Trajectory Synthesis (real-app).} The model generates long-tail interaction trajectories in data-scarce scenarios. This task tests whether generated states follow the requested interaction logic in less familiar settings.

    \item \textbf{Grounding-based Generation (grounding).} Given an initial screen and an action target specified by normalized coordinates in $[0,1000]$, the model generates the next GUI state. This task tests whether the output reflects an interaction with the intended control.

\end{enumerate}

We use the parenthetical task names throughout the results and analysis.

\begin{figure*}[htbp]
    \centering
    \includegraphics[width=\textwidth]{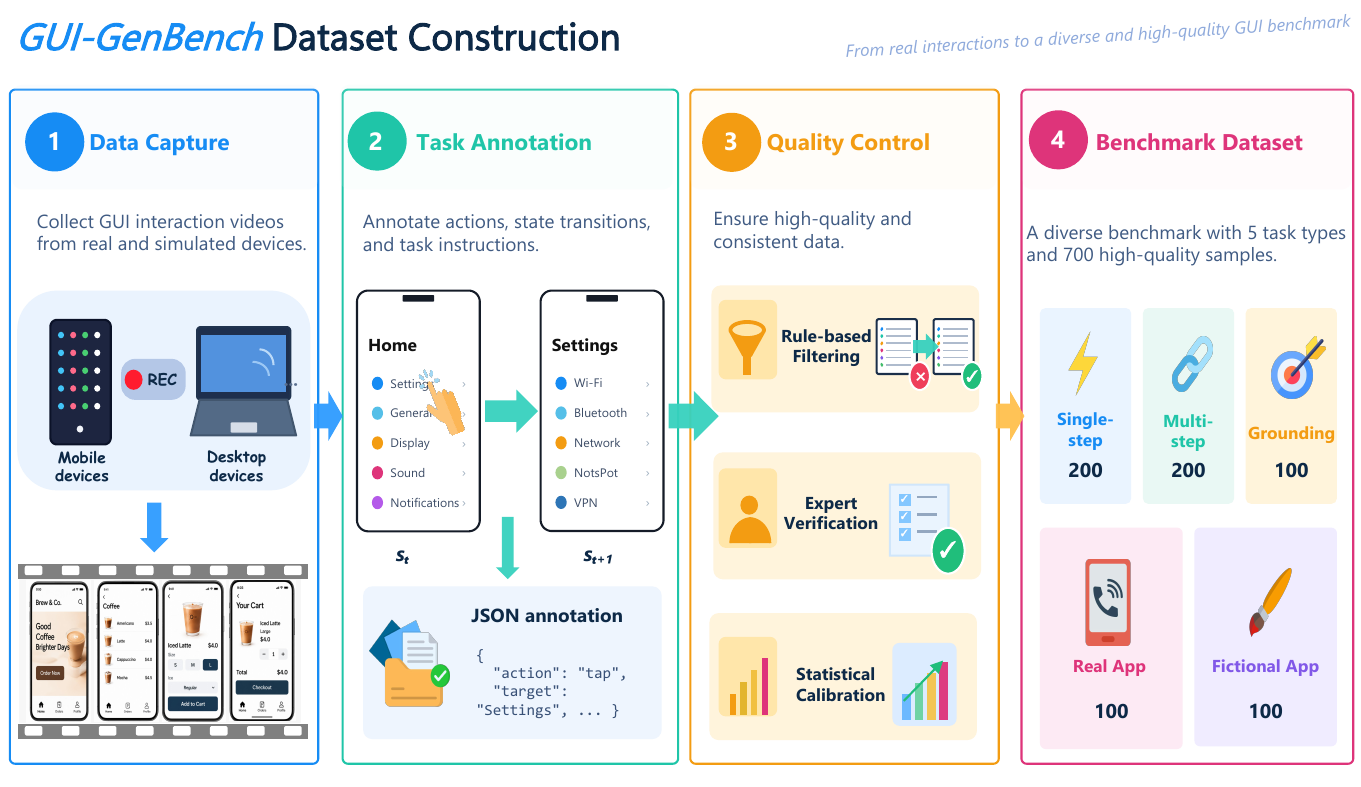}
    \caption{\textbf{GUI-GenBench data construction pipeline.} Interaction capture and task annotation are followed by preprocessing, human verification, and distribution calibration. The resulting 700 samples span the single-step, multi-step, grounding, real-app, and fictional-app task suites.}
    \label{fig:data_construction}
\end{figure*}

\subsection{Evaluation Dimensions and GUI-Score}
The five task suites specify \emph{what} a model must generate; five scoring dimensions assess \emph{how well} it does so. Task-specific rubrics evaluate functional correctness, consistency, and visual fidelity:

\begin{itemize}
    \item \textbf{Goal Achievement (GOAL)} assesses whether the generated state realizes the requested action or objective, including the correctness and completeness of the outcome.
    \item \textbf{Interaction Logic (LOGIC)} assesses whether state changes are consistent with valid actions and plausible interface behavior.
    \item \textbf{Content Consistency (CONS)} assesses preservation of content that should remain unchanged and stability of interface elements across a trajectory.
    \item \textbf{UI Plausibility (UI)} assesses structural coherence and conformity to interface conventions, including the absence of fabricated or incompatible controls.
    \item \textbf{Visual Quality (QUAL)} assesses text readability, icon clarity, and freedom from rendering artifacts.
\end{itemize}

Each dimension uses an ordinal scale from 0 to 5, with higher scores indicating better performance. Appendix~\ref{sec:rubrics} provides the task-specific criteria. For an evaluation set of $N$ samples, let $\mathcal{D}$ denote the five dimensions and $r_{i,d} \in \{0, \dots, 5\}$ the rating of sample $i$ on dimension $d$. GUI-Score averages the normalized ratings:

\begin{equation}
    \mathrm{GUI\text{-}Score} = \frac{1}{|\mathcal{D}| \cdot N} \sum_{d \in \mathcal{D}} \sum_{i=1}^{N} \left( \mathcal{F}(r_{i,d}) \right) = \frac{4}{N} \sum_{d=1}^{5} \sum_{i=1}^{N} r_{i,d}
\end{equation}

where $\mathcal{F}(r) = 20 \times r$ maps each rating to $[0,100]$. These values are normalized rubric scores, not percentages of successful interactions. We report dimension-level scores alongside the aggregate because visual quality can remain high when goal achievement is low. Appendix~\ref{sec:weight_sensitivity} compares equal weighting with a function-priority alternative; the rankings are similar under the two tested schemes.

\subsection{Data Construction Pipeline}
Figure~\ref{fig:data_construction} summarizes the construction pipeline. We collect screen recordings of mobile and desktop interactions. Annotators identify actions, such as clicking icons or scrolling, and represent the corresponding sequences as structured JSON metadata.

Quality control proceeds in three stages. Rule-based preprocessing filters inconsistent or noisy samples; human experts then check that annotated actions match the observed transitions. A final distribution-calibration step adjusts the sample mix, yielding 700 samples across the five task suites in Section~\ref{sec:tasks}. The Chinese and English subsets are sampled independently, not translated from a shared set of tasks. Screenshots for the real-app suite come from publicly released datasets with explicit licenses. Rare Trajectory Synthesis additionally uses fictional applications or uncommon scenarios expressed as text prompts, with the aim of reducing reliance on familiar screenshots. This design does not rule out overlap with model training data; Appendix~\ref{sec:limitations} discusses provenance and coverage limitations.

\begin{table*}[t]
    \centering
    \caption{\textbf{Main evaluation results on GUI-GenBench across Chinese and English subsets.} Scores are reported for five task suites and the aggregate GUI-Score. Within each language subset, \textbf{bold} and \underline{underlined} values indicate the best and second-best results in each column across all 12 models, respectively.}
    \label{tab:main_results}
    \scriptsize
    \setlength{\tabcolsep}{2.2pt}
    \renewcommand{\arraystretch}{1.2}
    \begin{tabular*}{\textwidth}{@{\extracolsep{\fill}}lrrrrrr@{\hspace{8pt}}rrrrrr@{}}
        \toprule
        \textbf{Model} & \multicolumn{6}{c}{\textbf{Chinese subset}} & \multicolumn{6}{c}{\textbf{English subset}} \\
        \cmidrule(lr){2-7} \cmidrule(l){8-13}
        & \shortstack{Single-\\step} & \shortstack{Multi-\\step} & \shortstack{Fictional-\\app} & \shortstack{Real-\\app} & \shortstack{Ground-\\ing} & \shortstack{\textbf{GUI-}\\\textbf{Score}} & \shortstack{Single-\\step} & \shortstack{Multi-\\step} & \shortstack{Fictional-\\app} & \shortstack{Real-\\app} & \shortstack{Ground-\\ing} & \shortstack{\textbf{GUI-}\\\textbf{Score}} \\
        \midrule
        \multicolumn{13}{@{}l}{\textit{Commercial models}} \\
        \addlinespace[2pt]
        Nano Banana pro & \textbf{84.50} & \textbf{68.65} & \textbf{65.75} & \underline{64.35} & \textbf{64.83} & \textbf{69.62} & \textbf{84.32} & \textbf{69.51} & 46.33 & 47.20 & \textbf{58.64} & \underline{61.20} \\
        Nano Banana & 64.36 & 34.16 & \underline{64.82} & \textbf{65.89} & 54.48 & 56.74 & 64.80 & 50.75 & 48.88 & 47.12 & 49.04 & 52.12 \\
        GPT-image-1.5 & \underline{83.79} & \underline{56.97} & 60.11 & 55.65 & 53.33 & \underline{63.22} & \underline{80.80} & 58.87 & \textbf{63.68} & \textbf{58.93} & 49.23 & \textbf{63.16} \\
        GPT-image-1.0 & 64.72 & 49.20 & 57.31 & 59.04 & 31.68 & 52.39 & 60.92 & \underline{64.33} & \underline{58.94} & \underline{56.16} & 37.84 & 55.64 \\
        Seedream 4.5 & 63.64 & 53.11 & 56.48 & 53.44 & 52.90 & 55.91 & 49.49 & 45.30 & 53.81 & 51.80 & 49.63 & 50.01 \\
        Seedream 4.0 & 62.04 & 48.64 & 49.28 & 50.93 & 53.53 & 52.88 & 53.28 & 37.57 & 47.92 & 49.36 & 44.17 & 46.46 \\
        Wan 2.6 & 64.20 & 50.11 & 52.72 & 50.40 & \underline{59.58} & 55.40 & 60.17 & 44.36 & 49.55 & 44.80 & 53.36 & 50.45 \\
        Flux-2-pro & 68.83 & 55.07 & 58.13 & 55.41 & 50.24 & 57.54 & 61.00 & 52.17 & 49.92 & 47.16 & 45.67 & 51.18 \\
        \midrule
        \multicolumn{13}{@{}l}{\textit{Open-source models}} \\
        \addlinespace[2pt]
        Bagel & 34.84 & 13.45 & 27.36 & 33.52 & 35.10 & 28.85 & 32.91 & 8.61 & 26.08 & 35.12 & 37.30 & 28.00 \\
        UniWorld-V2 & 55.33 & 24.95 & 32.03 & 21.39 & 49.60 & 36.66 & 42.68 & 14.14 & 30.08 & 26.83 & 47.04 & 32.15 \\
        Qwen-Image-Edit & 41.12 & 26.79 & 23.78 & 26.10 & 50.80 & 33.72 & 40.12 & 18.61 & 25.80 & 25.95 & \underline{54.55} & 33.01 \\
        Longcat-Image & 48.76 & 12.75 & 30.03 & 30.00 & 51.02 & 34.51 & 36.69 & 8.44 & 37.30 & 36.83 & 47.12 & 33.28 \\
        \bottomrule
    \end{tabular*}
\end{table*}

\Needspace{6\baselineskip}
\section{Evaluation}
\label{sec:evaluation}

\subsection{Evaluation Setup}

\paragraph{Evaluated models.} We evaluate 12 image generation models: 8 commercial and 4 open-source models.

The commercial models are Nano Banana Pro~\citep{team2023gemini}, Nano Banana~\citep{team2023gemini}, GPT-Image-1.5~\citep{openai2025gptimage}, GPT-Image-1~\citep{openai2025gptimage}, Seedream~4.5~\citep{seedream2025seed}, Seedream~4.0~\citep{seedream2025seedream}, Wan~2.6~\citep{wan2025}, and Flux-2-Pro~\citep{flux-2-2025}.

The open-source models are Bagel~\citep{deng2025emerging}, UniWorld-V2~\citep{li2025uniworld}, Qwen-Image-Edit~\citep{wu2025qwen}, and LongCat-Image~\citep{team2025longcat}.

\paragraph{VLM-based judges.} We evaluate generated outputs with three VLMs: Gemini-3-Pro-Native~\citep{google2025gemini3}, GPT-4o~\citep{hurst2024gpt}, and Qwen3-VL-235B-A22B-Thinking~\citep{bai2025qwen3vltechnicalreport}. We use official default configurations and evaluate each generated trajectory three times. Judge-specific results in Appendix~\ref{sec:judge_results} expose variation across evaluators; using multiple judges does not eliminate shared evaluation biases.

\paragraph{Multi-step generation protocol.} Given the initial screen and a target instruction, the model generates the subsequent five frames in one pass. No intermediate ground-truth frames are supplied. The task therefore evaluates a jointly generated trajectory rather than a sequence of edits conditioned on correct preceding states. It also differs from a closed-loop rollout in which an external agent selects a new action after each generated frame.

\begin{figure*}[t]
    \centering
    \includegraphics[height=0.45\textwidth]{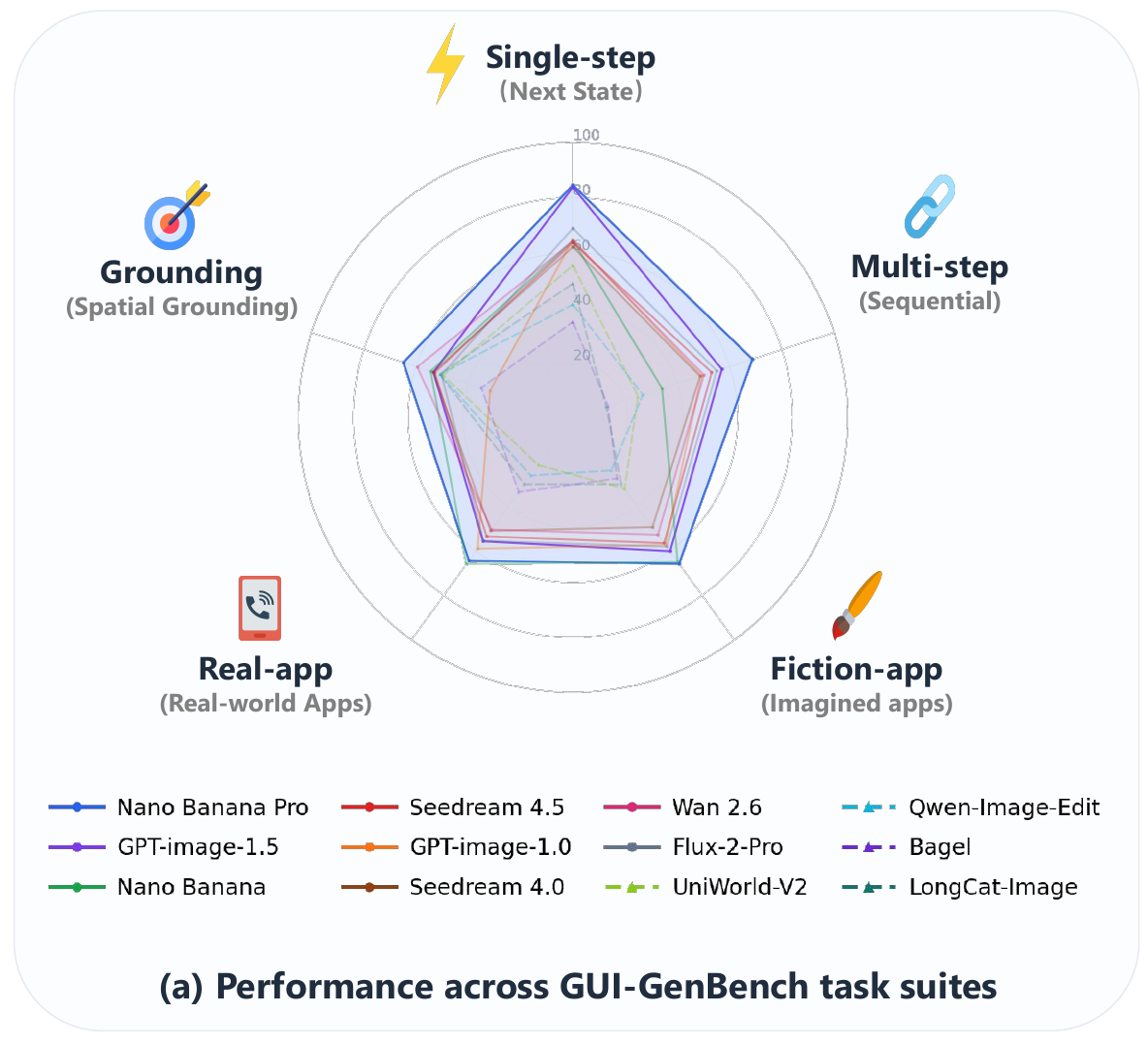}
    \hfill
    \includegraphics[height=0.45\textwidth]{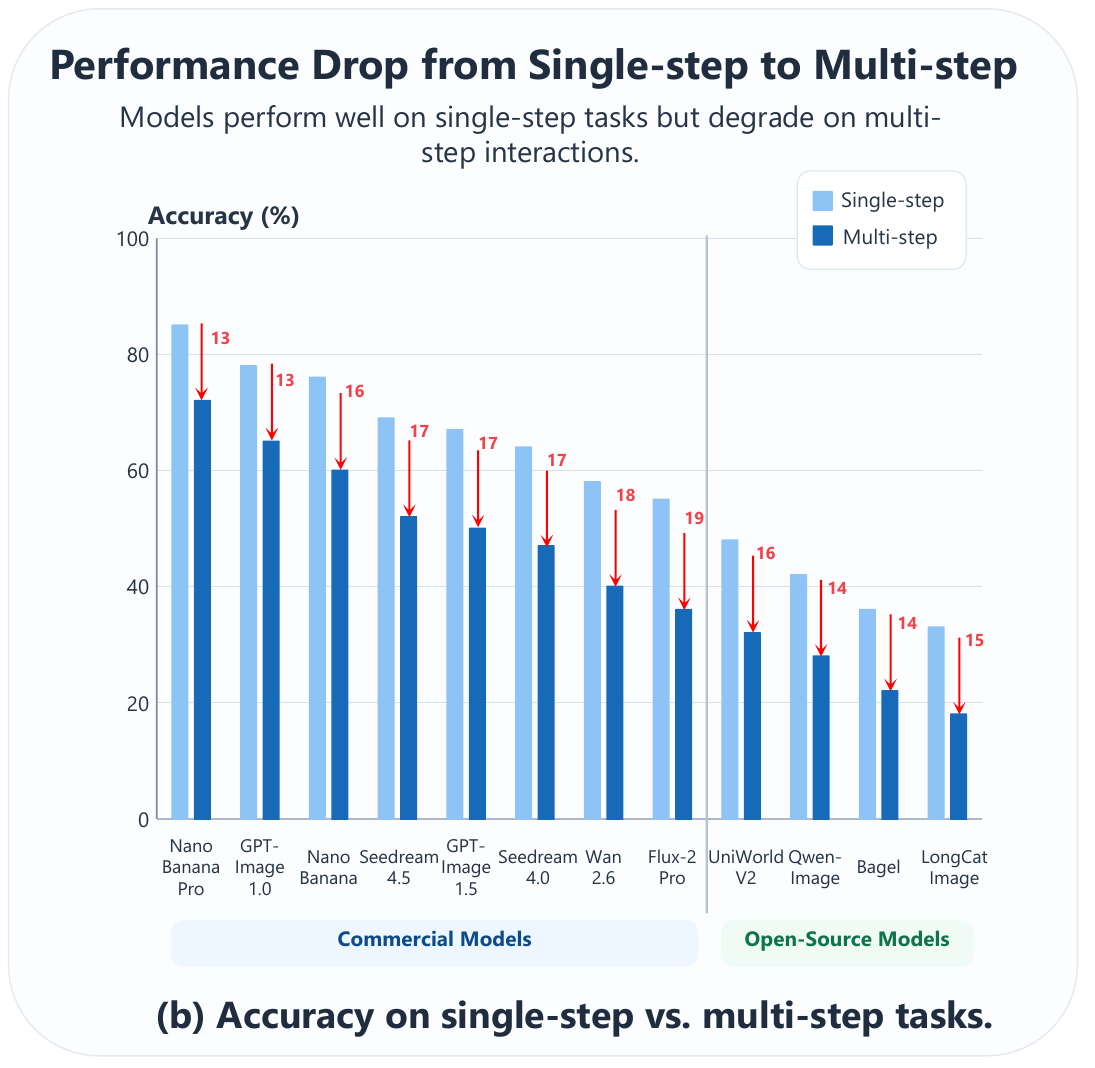}
    \caption{\textbf{Benchmark performance and single-step versus multi-step comparison.} Left: performance across the five task suites on the Chinese subset, with commercial models shown by solid lines and open-source models by dashed lines. Right: single-step and multi-step performance, with red arrows indicating the differences.}
    \label{fig:radar}
\end{figure*}

\subsection{Evaluation Results}

\paragraph{Overall performance.}
Table~\ref{tab:main_results} reports results separately for the Chinese and English subsets. Nano Banana Pro~\citep{team2023gemini} leads on the Chinese subset with a GUI-Score of 69.62, whereas GPT-Image-1.5~\citep{openai2025gptimage} leads on the English subset with 63.16. Commercial models generally outperform the evaluated open-source models, including UniWorld-V2~\citep{li2025uniworld} and Bagel~\citep{deng2025emerging}, although the size of the gap varies by task. Figure~\ref{fig:radar} summarizes the task-level profiles. Because the two language subsets contain independently sampled tasks, their score differences do not isolate the effect of language.

\paragraph{Single-step versus multi-step generation.}
Strong single-step scores do not consistently transfer to multi-step planning. Nano Banana Pro~\citep{team2023gemini} and GPT-Image-1.5~\citep{openai2025gptimage} both exceed 80 points on single-step tasks in each language subset, but receive lower multi-step scores. Across models, multi-step scores are often below 60 and, for some open-source models, below 10. The trend is not universal: GPT-Image-1 scores 64.33 on English multi-step tasks versus 60.92 on single-step tasks.

Multi-step generation requires both plausible individual screens and coherent dependencies between them. A trajectory can therefore fail despite visually convincing frames. However, the task suites differ in more than trajectory length, and all five frames are generated in one pass. The observed score gap should not be interpreted as a controlled estimate of error accumulation or as direct evidence about the models' internal reasoning mechanisms.

\paragraph{Coordinate grounding.}
Dimension-level scores reveal a weakness that aggregate scores can obscure. Under Gemini-3-Pro-Native, Nano Banana Pro~\citep{team2023gemini} obtains the highest grounding GOAL score, approximately 23.9 on the normalized 0--100 scale, while all four open-source models, including Qwen-Image-Edit~\citep{wu2025qwen}, score below 10 (Appendix Table~\ref{tab:gemini_full_metrics}). These are rubric scores, not action-success rates. By contrast, consistency and visual-quality scores can remain much higher, indicating that plausible appearance does not ensure a correct response to the coordinate-specified action. Appendix Figure~\ref{fig:type5_score} reports the grounding GOAL comparison, and Figure~\ref{fig:analysis} illustrates an incorrect transition despite a recognizable interface.

\subsection{Human--VLM Score Correlation}
We compare VLM scores with human judgments for five generators spanning a broad performance range: Nano Banana Pro~\citep{team2023gemini}, GPT-Image-1~\citep{openai2025gptimage}, Qwen-Image-Edit~\citep{wu2025qwen}, Bagel~\citep{deng2025emerging}, and LongCat-Image~\citep{team2025longcat}. For each generator, we select 10 samples from each of the 10 language--task combinations, giving 100 samples per generator. Four human experts independently apply the same scoring criteria as the VLM judges. We average their ratings before computing Pearson correlation with the VLM scores.

The initial comparison for Nano Banana Pro~\citep{team2023gemini} and GPT-Image-1~\citep{openai2025gptimage} yields $r=0.9926$ and $r=0.9833$, respectively (Appendix Figure~\ref{fig:humanscore}). Extending the comparison to lower-scoring generators gives $r=0.9647$ for Qwen-Image-Edit, $r=0.9613$ for Bagel, and $r=0.9577$ for LongCat-Image. The mean across the five correlations is $0.9719$, with a minimum of $0.9577$. Thus, high correlation is not confined to the two strongest generators in this study. Correlation measures association, however, not absolute score agreement, and does not establish that judges detect every interface error. Appendix~\ref{sec:human_validation} gives the calculation and limitations.

\subsection{Sensitivity to Scoring Choices}

\paragraph{Dimension weights.} Equal weighting treats functional and visual dimensions symmetrically. To examine whether emphasizing function changes the model comparison, we use the alternative weights reported in Appendix~\ref{sec:weight_sensitivity}: $30\%$ for GOAL, $30\%$ for LOGIC, $20\%$ for CONS, and $10\%$ each for UI and QUAL. Under Gemini-3-Pro-Native, the resulting ranking of the 12 models has Spearman correlation $\rho=0.979$ ($p<10^{-7}$) with the equal-weight ranking. The eight leading models retain their order, and only adjacent positions change. Thus, the leading-model comparison is unchanged under this particular increase in functional emphasis. The result does not imply invariance to other weight choices or judges.

\paragraph{Judge variation.} The detailed tables also show why judge identity matters. For Nano Banana Pro, the multi-step Overall score is 69.07 under Gemini-3-Pro-Native, 87.68 under GPT-4o, and 91.71 under Qwen3-VL-235B-A22B-Thinking (Appendix Tables~\ref{tab:gemini_full_metrics}--\ref{tab:model_evaluation_new}). Mapping ratings to a common 0--100 range does not remove these differences. We therefore retain the judge-specific results alongside the task-level summary, so that readers can distinguish a generator comparison under a given evaluator from changes attributable to the evaluator. Correlation with human ratings and sensitivity to dimension weights address different aspects of evaluation; neither replaces inspection of judge-dependent scores.

\section{Discussion and Analysis}
\label{sec:discussion}

\subsection{From Plausible Screens to Reliable Transitions}

Table~\ref{tab:main_results} and Figure~\ref{fig:radar} suggest that local transition quality is an incomplete measure of a generative GUI environment. Leading models~\citep{team2023gemini, openai2025gptimage} exceed 80 points on single-step tasks, yet show weaker performance on multi-step trajectories. The grounding results expose a related distinction between preserving appearance and executing the intended action. Together, these observations motivate inspecting functional dimensions alongside aggregate scores. They do not establish a universal ordering of task difficulty or show that a model relies on memorization rather than interaction dynamics.

GUI-GenBench evaluates generated visual states, not executable software or downstream agent learning. Its diagnostic value is to reveal where a plausible screen ceases to be a faithful response. Whether filtering synthetic trajectories using these diagnostics improves agent training remains an open empirical question.

% Let the report layout flow naturally; do not retain ICLR page-break tuning.
\Needspace{8\baselineskip}
\subsection{Qualitative Analysis of Failure Cases}

\begin{figure}[H]
    \centering
    \includegraphics[width=0.90\textwidth]{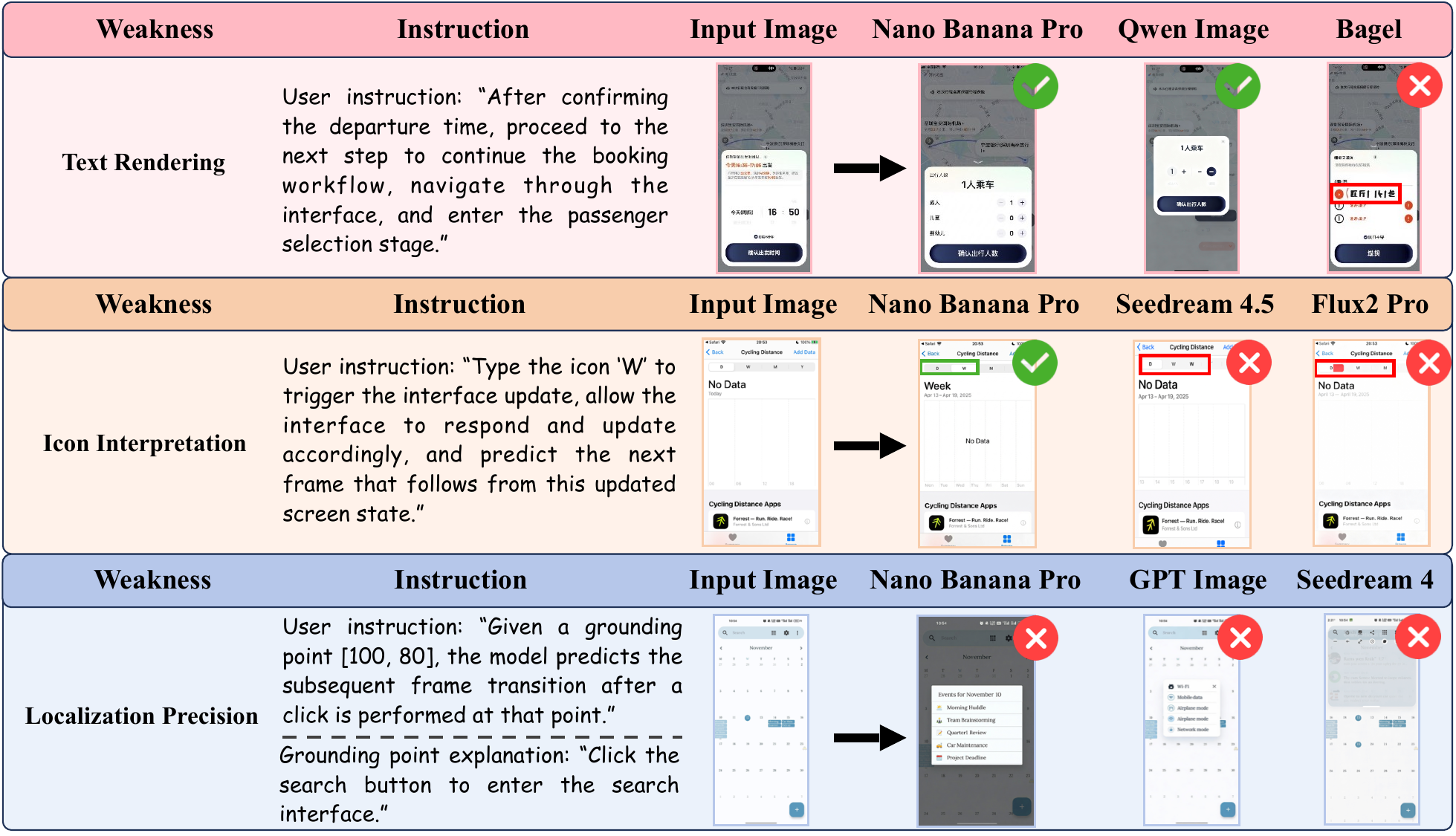}
    \caption{\textbf{Representative failure patterns in generated GUI states.} Top: text rendering in a passenger-selection interface. Middle: interpreting the ``W'' control to switch to a weekly view. Bottom: executing a coordinate-based instruction targeting the search control. Green checks and red crosses indicate successful and failed responses in the shown examples.}
    \label{fig:analysis}
\end{figure}

Figure~\ref{fig:analysis} pairs instructions and reference screens with generated outputs to illustrate three failure modes. The examples identify specific errors; they do not estimate how frequently those errors occur across the benchmark.

\textbf{Text rendering (top row).} The booking task requires a transition to passenger selection while preserving readable text. Nano Banana Pro~\citep{team2023gemini} and the Qwen Image output are marked as successful; Bagel produces corrupted characters in the highlighted region. A recognizable layout therefore does not guarantee readable content, and the failure is not shared by every model in this example.

\textbf{Icon interpretation (middle row).} The instruction targets the ``W'' control on a cycling-distance screen. Nano Banana Pro generates the weekly view, whereas the highlighted Seedream 4.5 and Flux-2-Pro outputs fail to realize that transition. The required change depends on the control's functional meaning, not merely its visual appearance.

\textbf{Coordinate grounding (bottom row).} The point $[100,80]$ targets the search control, but the generated outputs show unrelated menus or overlays. This mismatch between the requested target and the resulting state complements the low grounding GOAL scores in Section~\ref{sec:evaluation}. A visually plausible screen remains functionally incorrect when it depicts a different interaction.

\Needspace{7\baselineskip}
\subsection*{AI use statement}
Generative AI tools assisted with \LaTeX{} template migration,
manuscript organization and language editing, figure-layout drafts, and
refinement of the written interpretation of existing results. VLMs also serve
as evaluators in the benchmark, as described in Section~\ref{sec:evaluation}.
The authors are responsible for verifying the AI-assisted material and for
the final text, claims, and artifacts.
% Before submission, authors must verify this disclosure against all actual
% AI uses and complete human review of the final manuscript and artifacts.

\FloatBarrier
\bibliographystyle{unsrtnat}
\bibliography{GEBench}

\clearpage

\setcounter{figure}{0}
\makeatletter
\renewcommand{\thefigure}{A\@arabic\c@figure}
\renewcommand{\theHfigure}{appendix.\arabic{figure}}
\renewcommand{\p@subfigure}{\thefigure}
\setlength{\@fptop}{0pt}
\setlength{\@dblfptop}{0pt}
\makeatother

\setcounter{table}{0}
\makeatletter
\renewcommand{\thetable}{A\@arabic\c@table}
\renewcommand{\theHtable}{appendix.\arabic{table}}
\makeatother

\appendix

\section*{Appendix}

\section{GUI-Score Weighting Sensitivity}
\label{sec:weight_sensitivity}

GUI-Score gives equal weight to Goal Achievement, Interaction Logic, Content Consistency, UI Plausibility, and Visual Quality. To examine sensitivity to this choice, we compare it with a function-priority scheme using weights of $30\%$, $30\%$, $20\%$, $10\%$, and $10\%$, respectively. We re-rank all 12 models using the Gemini-3-Pro-Native scores under each scheme. The rankings have a Spearman correlation of $\rho=0.979$ ($p<10^{-7}$); only adjacent positions exchange order, and the eight leading models retain their ordering. This comparison supports ranking stability for the two tested schemes under this judge, not robustness to arbitrary weights. Dimension-level scores remain important when functional correctness is the primary concern.

\section{Extended Human Validation}
\label{sec:human_validation}

The human study uses 100 samples per generator, with 10 drawn from each language--task combination and scored independently by four experts. For each generator, let $h_i$ denote a human score averaged across the four experts and $v_i$ the corresponding VLM score; $\bar h$ and $\bar v$ denote their respective means. We quantify their linear association using Pearson correlation:
\begin{equation}
 r = \frac{\sum_i (h_i-\bar{h})(v_i-\bar{v})}
 {\sqrt{\sum_i(h_i-\bar{h})^2}\sqrt{\sum_i(v_i-\bar{v})^2}}.
\end{equation}
The correlations are $0.9926$ for Nano Banana Pro, $0.9833$ for GPT-Image-1, $0.9647$ for Qwen-Image-Edit, $0.9613$ for Bagel, and $0.9577$ for LongCat-Image, with a mean of $0.9719$. These results show a strong linear association in the reported comparisons across generators of different quality. They do not establish absolute score calibration, agreement among human annotators, or detection of every fine-grained interface error. Inter-rater agreement and unanimous-case frequency would provide complementary evidence from the per-annotator labels. The reported correlations should therefore be interpreted as one check on the scoring framework, not a complete validation of judge reliability.

\begin{figure}[H]
    \centering
    \subcaptionbox{Human--VLM score correlation.\label{fig:humanscore}}{
        \includegraphics[width=0.38\textwidth]{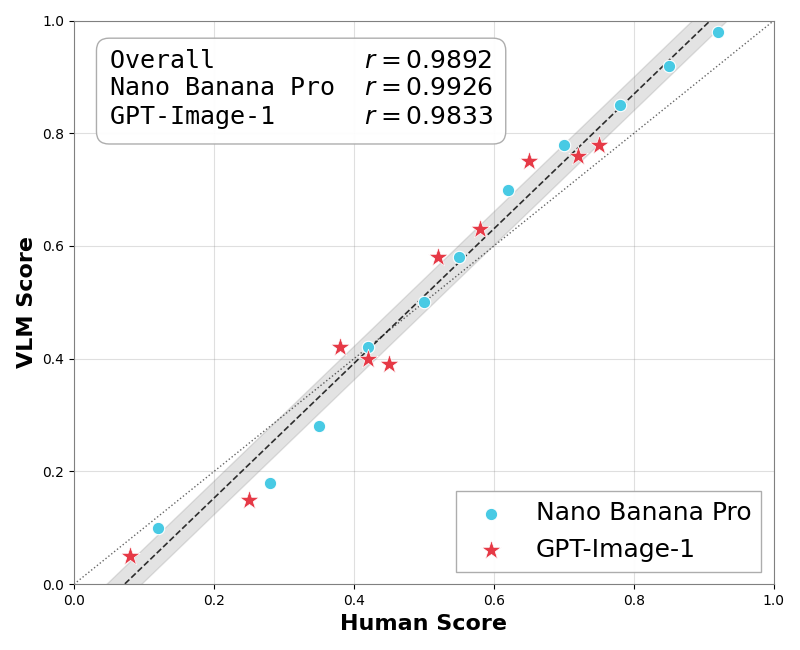}}
    \hfill
    \subcaptionbox{GOAL scores on the grounding task.\label{fig:type5_score}}{
        \includegraphics[width=0.57\textwidth]{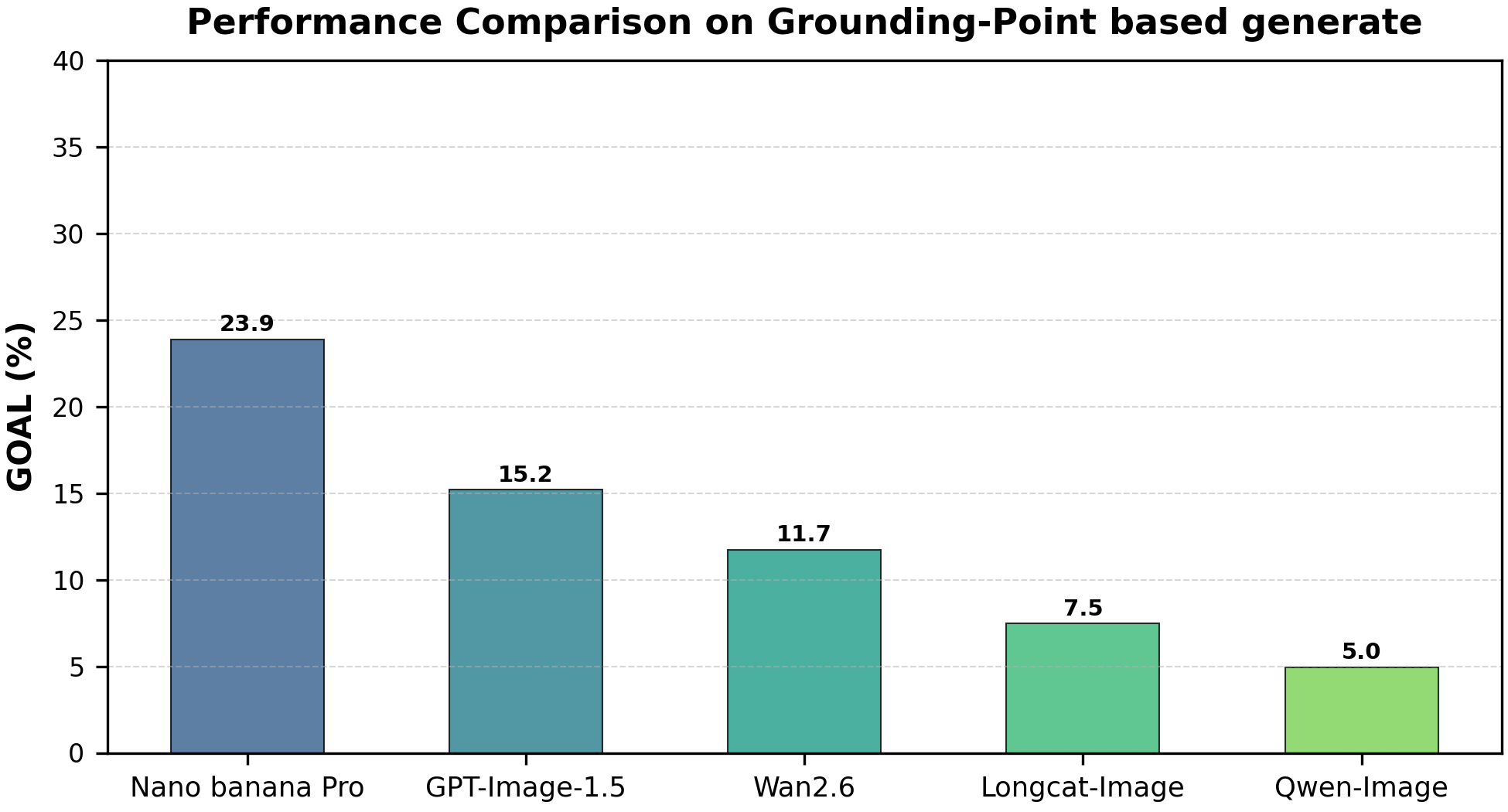}}
    \caption{\textbf{Human--VLM correlation and grounding performance.} Left: the initial score-correlation comparison for Nano Banana Pro~\citep{team2023gemini} and GPT-Image-1~\citep{openai2025gptimage}; the text reports the extended five-model results. Right: normalized grounding GOAL scores, which assess whether generated states reflect the coordinate-specified action rather than simply preserving visual appearance.}
    \label{fig:validation_grounding}
\end{figure}

\section{Benchmark Scope, Data Provenance, and Limitations}
\label{sec:limitations}

\paragraph{Data provenance and language subsets.}
The Chinese and English subsets contain independently sampled tasks rather than paired translations. Cross-language score differences can therefore reflect differences in task composition as well as language. Screenshots in the real-app suite come from publicly released datasets with explicit licenses; any benchmark release should include a per-sample source and license manifest. Rare Trajectory Synthesis also uses fictional applications and uncommon scenarios described through text prompts. These scenarios are intended to reduce dependence on familiar screenshots, but we cannot exclude training-data overlap for the evaluated generators.

\paragraph{Coverage and temporal validity.}
The 700 samples do not exhaust the diversity of applications, platforms, languages, or actions, and do not cover all emerging interfaces, including augmented- and virtual-reality systems. The sampled application and interaction distributions may introduce domain bias. Interface redesigns and software updates can also make captured states outdated. Results thus describe a benchmark snapshot rather than all GUI interactions. Future releases should record capture time, application version where available, source, platform, language, action type, and license.

\paragraph{Evaluation limitations.}
High human--VLM correlation does not exclude shared judge biases or missed interface errors. Normalizing all judges to the same numerical range also does not make their scores equally calibrated. We do not use CLIP-family similarity as a primary metric: small text, icon, or button-state errors can matter functionally despite high global similarity. Localization overlap metrics such as IoU are relevant where spatial annotations exist, but do not capture whole-interface correctness. Such metrics and human--human agreement would provide complementary evidence. Finally, our single-shot trajectory protocol does not test closed-loop agent interaction or the effect of generated data on downstream training.

\paragraph{Safety implications.}
Trajectories containing fabricated controls, incorrect transitions, or grounding errors should not be treated as faithful execution traces. Training agents on such outputs may reinforce invalid or unsafe behavior. GUI-GenBench provides a diagnostic of generated content, not a guarantee of deployment safety. Synthetic trajectories require filtering and independent validation before use in agent training or safety-critical interaction.

\section{Evaluation Framework}
\begin{figure*}[htbp]
    \centering
    \includegraphics[width=1\textwidth]{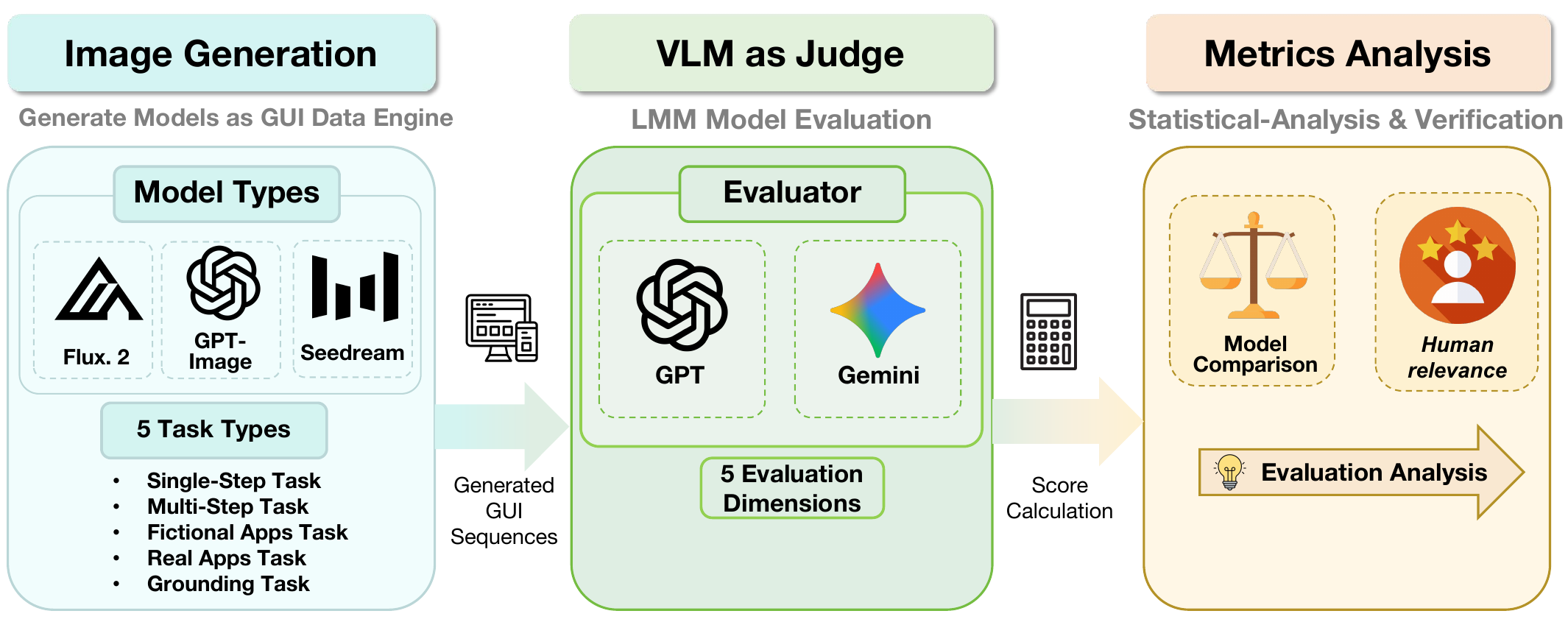}
    \caption{\textbf{GUI-GenBench evaluation framework.} Generated screens and trajectories are scored by three VLM judges using task-specific rubrics. Analysis summarizes task- and dimension-level scores and compares VLM scores with human ratings.}
    \label{fig:evaluation_framework}
\end{figure*}

Figure~\ref{fig:evaluation_framework} summarizes three stages. \textbf{Generation} produces a screen or trajectory from the task instruction and, where specified, a reference state. \textbf{Scoring} applies the task-specific rubrics to GOAL, LOGIC, CONS, UI, and QUAL using GPT-4o~\citep{hurst2024gpt}, Gemini-3-Pro-Native~\citep{google2025gemini3}, and Qwen3-VL-235B-A22B-Thinking~\citep{bai2025qwen3vltechnicalreport}. \textbf{Analysis} summarizes scores by task and dimension, compares the judges, and examines correlation with human ratings. Keeping these stages distinct separates generation failures from variation in how evaluators score them.

\FloatBarrier
\Needspace{10\baselineskip}
\section{Detailed Performance on GUI-GenBench Using Different Judges}
\label{sec:judge_results}
The following tables report all 12 generators under each VLM judge, allowing model comparisons to be examined separately for each evaluator:

\begin{itemize}
    \item Table~\ref{tab:gemini_full_metrics}: Detailed scores assigned by Gemini-3-Pro-Native~\citep{google2025gemini3} as the judge.
    \item Table~\ref{tab:gpt4o_full_metrics}: Detailed scores assigned by GPT-4o~\citep{hurst2024gpt} as the judge.
    \item Table~\ref{tab:model_evaluation_new}: Detailed scores assigned by Qwen3-VL-235B-A22B-Thinking~\citep{bai2025qwen3vltechnicalreport} as the judge.
\end{itemize}

For every task suite, each table reports GOAL, LOGIC, CONS, UI, QUAL, and an Overall score. Raw ratings are linearly mapped to $[0,100]$ before aggregation. This common reporting range does not imply identical calibration across judges; the separate tables expose evaluator-dependent variation rather than treating any one judge as ground truth.

\begin{table*}[!htbp]
\centering
\caption{Detailed GUI-Score results on GUI-GenBench with \textbf{Gemini-3-Pro-Native}~\citep{google2025gemini3} as the judge.}
\label{tab:gemini_full_metrics}
\setlength{\tabcolsep}{1.5pt}
\renewcommand{\arraystretch}{0.9}
\resizebox{\textwidth}{!}{%
\begin{tabular}{ll|cccccccccccc}
\toprule
\textbf{Task} & \textbf{Metric} & 
\begin{tabular}[c]{@{}c@{}}\textbf{Nano}\\\textbf{Banana Pro}\end{tabular} & 
\begin{tabular}[c]{@{}c@{}}\textbf{Nano}\\\textbf{Banana}\end{tabular} & 
\begin{tabular}[c]{@{}c@{}}\textbf{GPT-Image}\\\textbf{-1.5}\end{tabular} & 
\begin{tabular}[c]{@{}c@{}}\textbf{GPT-Image}\\\textbf{-1.0}\end{tabular} & 
\begin{tabular}[c]{@{}c@{}}\textbf{Seedream}\\\textbf{4.5}\end{tabular} & 
\begin{tabular}[c]{@{}c@{}}\textbf{Seedream}\\\textbf{4.0}\end{tabular} & 
\begin{tabular}[c]{@{}c@{}}\textbf{Wan 2.6}\end{tabular} & 
\begin{tabular}[c]{@{}c@{}}\textbf{Flux-2-Pro}\end{tabular} & 
\begin{tabular}[c]{@{}c@{}}\textbf{Bagel}\end{tabular} & 
\begin{tabular}[c]{@{}c@{}}\textbf{UniWorld}\end{tabular} & 
\begin{tabular}[c]{@{}c@{}}\textbf{Qwen}\\\textbf{-Image-Edit}\end{tabular} & 
\begin{tabular}[c]{@{}c@{}}\textbf{Longcat}\\\textbf{-Image}\end{tabular} \\ \midrule

\multirow{6}{*}{\centering\shortstack{single-step}} 
 & GOAL & \cellcolor{firstcolor}\textbf{95.04} & 69.60 & \cellcolor{secondcolor}87.65 & \cellcolor{thirdcolor}86.50 & 56.60 & 54.58 & 58.94 & 65.08 & 24.35 & 41.00 & 25.60 & 36.07 \\
 & LOGIC & \cellcolor{secondcolor}82.04 & 58.50 & \cellcolor{firstcolor}\textbf{86.50} & \cellcolor{thirdcolor}78.80 & 49.25 & 45.29 & 52.17 & 53.60 & 29.12 & 43.33 & 27.93 & 35.47 \\
 & CONS & \cellcolor{thirdcolor}75.24 & 69.70 & 72.65 & 58.10 & 66.23 & 65.68 & \cellcolor{firstcolor}\textbf{81.81} & \cellcolor{secondcolor}77.46 & 54.76 & 58.77 & 54.93 & 53.73 \\
 & UI & \cellcolor{secondcolor}79.50 & 51.60 & \cellcolor{firstcolor}\textbf{84.76} & \cellcolor{thirdcolor}58.00 & 50.19 & 47.87 & 53.50 & 48.43 & 28.44 & 47.93 & 40.07 & 42.60 \\
 & QUAL & \cellcolor{firstcolor}\textbf{90.22} & 73.50 & \cellcolor{secondcolor}79.97 & 32.00 & 61.89 & 68.39 & 64.65 & \cellcolor{thirdcolor}79.70 & 30.95 & 54.00 & 52.00 & 45.73 \\
 & Overall & \cellcolor{firstcolor}\textbf{84.41} & 64.58 & \cellcolor{secondcolor}82.31 & 62.68 & 56.83 & 56.36 & 62.21 & \cellcolor{thirdcolor}64.85 & 33.52 & 49.01 & 40.11 & 42.72 \\ \midrule

\multirow{6}{*}{\centering\shortstack{multi-step}} 
 & GOAL & \cellcolor{firstcolor}\textbf{86.63} & 47.61 & \cellcolor{secondcolor}79.13 & \cellcolor{thirdcolor}77.45 & 68.25 & 60.14 & 46.89 & 67.18 & 3.85 & 2.29 & 1.98 & 1.78 \\
 & LOGIC & \cellcolor{firstcolor}\textbf{56.75} & 30.46 & 36.94 & 35.31 & 36.63 & \cellcolor{thirdcolor}38.59 & 25.54 & \cellcolor{secondcolor}52.71 & 8.87 & 3.73 & 6.70 & 2.96 \\
 & CONS & \cellcolor{firstcolor}\textbf{56.95} & 45.69 & 36.39 & 34.08 & 38.66 & 38.70 & \cellcolor{secondcolor}56.43 & \cellcolor{thirdcolor}46.46 & 15.46 & 31.86 & 43.32 & 19.89 \\
 & UI & \cellcolor{firstcolor}\textbf{54.96} & 26.80 & \cellcolor{secondcolor}50.93 & \cellcolor{thirdcolor}50.41 & 35.15 & 33.44 & 37.56 & 29.79 & 12.10 & 20.17 & 29.17 & 16.08 \\
 & QUAL & \cellcolor{firstcolor}\textbf{90.08} & 62.03 & \cellcolor{secondcolor}86.23 & \cellcolor{thirdcolor}85.82 & 66.80 & 48.04 & 64.88 & 63.18 & 13.54 & 40.41 & 31.60 & 10.41 \\
 & Overall & \cellcolor{firstcolor}\textbf{69.07} & 42.52 & \cellcolor{secondcolor}57.92 & \cellcolor{thirdcolor}56.61 & 49.10 & 43.78 & 46.26 & 51.86 & 10.76 & 19.69 & 22.55 & 10.22 \\ \midrule

\multirow{6}{*}{\centering\shortstack{fictional-app}} 
 & GOAL & \cellcolor{thirdcolor}40.51 & 40.40 & \cellcolor{firstcolor}\textbf{46.20} & 38.78 & 40.00 & 39.33 & 36.49 & \cellcolor{secondcolor}45.93 & 15.80 & 4.00 & 0.40 & 17.01 \\
 & LOGIC & \cellcolor{firstcolor}\textbf{51.03} & \cellcolor{thirdcolor}48.28 & 43.13 & 40.61 & 39.73 & 39.80 & 42.54 & \cellcolor{secondcolor}48.60 & 29.80 & 4.60 & 0.81 & 20.90 \\
 & CONS & 57.95 & 60.00 & 53.20 & 53.88 & 61.00 & \cellcolor{thirdcolor}64.68 & 61.17 & 50.67 & 41.20 & \cellcolor{firstcolor}\textbf{88.07} & 62.36 & \cellcolor{secondcolor}85.00 \\
 & UI & \cellcolor{thirdcolor}50.26 & 45.45 & \cellcolor{firstcolor}\textbf{68.72} & \cellcolor{secondcolor}61.63 & 42.60 & 39.40 & 41.24 & 39.00 & 27.60 & 28.03 & 27.95 & 25.42 \\
 & QUAL & 89.49 & 89.70 & \cellcolor{firstcolor}\textbf{98.20} & \cellcolor{secondcolor}95.71 & \cellcolor{thirdcolor}92.40 & 59.87 & 74.16 & 85.93 & 19.20 & 30.47 & 32.39 & 20.90 \\
 & Overall & \cellcolor{thirdcolor}57.85 & 56.77 & \cellcolor{firstcolor}\textbf{61.89} & \cellcolor{secondcolor}58.12 & 55.15 & 48.62 & 51.12 & 54.03 & 26.72 & 31.03 & 24.78 & 33.85 \\ \midrule

\multirow{6}{*}{\centering\shortstack{real-app}} 
 & GOAL & 28.73 & 32.80 & 26.96 & \cellcolor{thirdcolor}34.75 & \cellcolor{secondcolor}40.00 & \cellcolor{firstcolor}\textbf{43.80} & 33.16 & 34.15 & 18.20 & 2.06 & 0.40 & 17.07 \\
 & LOGIC & \cellcolor{thirdcolor}44.40 & \cellcolor{secondcolor}45.20 & 36.36 & 39.80 & 37.20 & 43.27 & 41.10 & \cellcolor{firstcolor}\textbf{47.24} & 33.80 & 2.06 & 0.40 & 20.00 \\
 & CONS & 67.13 & 65.33 & 56.83 & 56.77 & 55.00 & \cellcolor{thirdcolor}67.87 & 59.07 & 50.71 & 63.60 & \cellcolor{secondcolor}72.23 & 67.67 & \cellcolor{firstcolor}\textbf{82.79} \\
 & UI & \cellcolor{thirdcolor}48.00 & 46.80 & \cellcolor{firstcolor}\textbf{70.24} & \cellcolor{secondcolor}58.79 & 41.20 & 40.60 & 39.24 & 39.25 & 34.20 & 22.54 & 27.60 & 25.99 \\
 & QUAL & 90.60 & \cellcolor{thirdcolor}92.40 & \cellcolor{secondcolor}96.16 & \cellcolor{firstcolor}\textbf{97.84} & 89.80 & 55.20 & 68.78 & 84.25 & 21.80 & 21.79 & 32.80 & 20.88 \\
 & Overall & 55.77 & \cellcolor{thirdcolor}56.51 & \cellcolor{secondcolor}57.31 & \cellcolor{firstcolor}\textbf{57.59} & 52.64 & 50.15 & 48.27 & 51.12 & 34.32 & 24.14 & 25.77 & 33.35 \\ \midrule

\multirow{6}{*}{\centering\shortstack{grounding}} 
 & GOAL & \cellcolor{firstcolor}\textbf{23.88} & \cellcolor{secondcolor}21.80 & 15.19 & 15.51 & \cellcolor{thirdcolor}16.12 & 15.42 & 11.74 & 11.29 & 5.86 & 4.00 & 4.95 & 7.47 \\
 & LOGIC & 36.94 & \cellcolor{thirdcolor}41.00 & 37.39 & 37.55 & \cellcolor{firstcolor}\textbf{51.84} & \cellcolor{secondcolor}44.58 & 35.65 & 36.13 & 17.17 & 18.20 & 14.43 & 15.35 \\
 & CONS & \cellcolor{thirdcolor}77.96 & 71.20 & 51.20 & 45.31 & 64.29 & 67.71 & \cellcolor{firstcolor}\textbf{83.26} & 69.68 & 62.42 & 69.80 & \cellcolor{secondcolor}80.21 & 74.95 \\
 & UI & \cellcolor{secondcolor}77.35 & 52.00 & 66.90 & 61.77 & 48.57 & 48.96 & 68.04 & 50.32 & 41.62 & \cellcolor{thirdcolor}70.40 & \cellcolor{firstcolor}\textbf{80.41} & 74.34 \\
 & QUAL & \cellcolor{firstcolor}\textbf{92.24} & 72.80 & \cellcolor{secondcolor}85.57 & 81.50 & 75.51 & 67.08 & \cellcolor{thirdcolor}83.26 & 79.35 & 53.94 & 79.20 & 83.09 & 73.13 \\
 & Overall & \cellcolor{firstcolor}\textbf{61.67} & 51.76 & 51.25 & 48.33 & 51.27 & 48.75 & \cellcolor{secondcolor}56.39 & 49.35 & 36.20 & 48.32 & \cellcolor{thirdcolor}52.62 & 49.05 \\ \bottomrule

\end{tabular}%
}
\end{table*}

\begin{table*}[t]
\centering
\caption{Detailed GUI-Score results on GUI-GenBench with \textbf{GPT-4o}~\citep{hurst2024gpt} as the judge.}
\label{tab:gpt4o_full_metrics}

\setlength{\tabcolsep}{1.5pt}   % 原2pt
\renewcommand{\arraystretch}{0.9} % 行距压缩（非常关键）

\resizebox{\textwidth}{!}{%
\begin{tabular}{ll|cccccccccccc}
\toprule
\textbf{Task} & \textbf{Metric} & 
\begin{tabular}[c]{@{}c@{}}\textbf{Nano}\\\textbf{Banana Pro}\end{tabular} & 
\begin{tabular}[c]{@{}c@{}}\textbf{Nano}\\\textbf{Banana}\end{tabular} & 
\begin{tabular}[c]{@{}c@{}}\textbf{GPT-Image}\\\textbf{-1.5}\end{tabular} & 
\begin{tabular}[c]{@{}c@{}}\textbf{GPT-Image}\\\textbf{-1.0}\end{tabular} & 
\begin{tabular}[c]{@{}c@{}}\textbf{Seedream}\\\textbf{4.5}\end{tabular} & 
\begin{tabular}[c]{@{}c@{}}\textbf{Seedream}\\\textbf{4.0}\end{tabular} & 
\begin{tabular}[c]{@{}c@{}}\textbf{Wan 2.6}\end{tabular} & 
\begin{tabular}[c]{@{}c@{}}\textbf{Flux-2-Pro}\end{tabular} & 
\begin{tabular}[c]{@{}c@{}}\textbf{Bagel}\end{tabular} & 
\begin{tabular}[c]{@{}c@{}}\textbf{UniWorld}\end{tabular} & 
\begin{tabular}[c]{@{}c@{}}\textbf{Qwen}\\\textbf{-Image-Edit}\end{tabular} & 
\begin{tabular}[c]{@{}c@{}}\textbf{Longcat}\\\textbf{-Image}\end{tabular} \\ \midrule

\multirow{6}{*}{\centering\shortstack{single-step}} 
 & GOAL & \cellcolor{firstcolor}\textbf{90.73} & 74.57 & \cellcolor{secondcolor}88.37 & 80.41 & 63.37 & 60.56 & \cellcolor{thirdcolor}71.38 & 73.10 & 31.40 & 47.30 & 37.40 & 40.28 \\
 & LOGIC & \cellcolor{firstcolor}\textbf{94.25} & 81.10 & \cellcolor{secondcolor}92.04 & 84.39 & 68.91 & 66.92 & \cellcolor{thirdcolor}77.18 & 78.07 & 38.00 & 52.20 & 42.40 & 44.68 \\
 & CONS & \cellcolor{firstcolor}\textbf{84.52} & 79.49 & 78.27 & 68.88 & 70.40 & 70.32 & \cellcolor{secondcolor}84.34 & \cellcolor{thirdcolor}83.25 & 55.90 & 62.20 & 64.60 & 61.44 \\
 & UI & \cellcolor{firstcolor}\textbf{91.69} & 78.09 & \cellcolor{secondcolor}88.57 & 82.96 & 66.00 & 65.55 & \cellcolor{thirdcolor}75.84 & 78.48 & 42.20 & 55.40 & 55.60 & 51.00 \\
 & QUAL & \cellcolor{secondcolor}78.95 & 69.54 & \cellcolor{firstcolor}\textbf{78.98} & \cellcolor{thirdcolor}77.14 & 55.37 & 58.71 & 69.68 & 77.06 & 37.80 & 53.20 & 56.90 & 48.56 \\
 & Overall & \cellcolor{firstcolor}\textbf{88.03} & 76.56 & \cellcolor{secondcolor}85.25 & \cellcolor{thirdcolor}78.76 & 64.81 & 64.41 & 75.68 & 77.99 & 41.06 & 54.06 & 51.38 & 49.19 \\ \midrule

\multirow{6}{*}{\centering\shortstack{multi-step}} 
 & GOAL & \cellcolor{thirdcolor}89.83 & 72.30 & \cellcolor{firstcolor}\textbf{95.44} & \cellcolor{secondcolor}95.31 & 88.92 & 76.96 & 77.21 & 83.74 & 18.31 & 26.21 & 15.63 & 14.01 \\
 & LOGIC & \cellcolor{thirdcolor}84.99 & 72.80 & \cellcolor{secondcolor}87.16 & \cellcolor{firstcolor}\textbf{87.45} & 80.43 & 70.14 & 73.03 & 78.38 & 26.83 & 36.56 & 32.28 & 29.53 \\
 & CONS & \cellcolor{firstcolor}\textbf{83.12} & 77.70 & \cellcolor{secondcolor}82.77 & \cellcolor{thirdcolor}82.04 & 77.70 & 68.34 & 81.89 & 80.51 & 31.47 & 56.11 & 68.36 & 44.75 \\
 & UI & \cellcolor{firstcolor}\textbf{89.80} & 76.60 & \cellcolor{thirdcolor}89.16 & \cellcolor{firstcolor}\textbf{89.80} & 84.39 & 71.55 & 80.88 & 79.49 & 27.27 & 52.60 & 55.82 & 33.87 \\
 & QUAL & \cellcolor{thirdcolor}90.68 & 78.40 & \cellcolor{secondcolor}92.32 & \cellcolor{firstcolor}\textbf{93.98} & 85.83 & 64.68 & 79.02 & 76.97 & 25.37 & 56.11 & 42.07 & 21.31 \\
 & Overall & \cellcolor{thirdcolor}87.68 & 75.56 & \cellcolor{secondcolor}89.37 & \cellcolor{firstcolor}\textbf{89.72} & 83.45 & 70.33 & 78.41 & 79.82 & 25.85 & 45.52 & 42.83 & 28.69 \\ \midrule

\multirow{6}{*}{\centering\shortstack{fictional-app}} 
 & GOAL & 40.36 & 42.07 & \cellcolor{firstcolor}\textbf{51.60} & \cellcolor{secondcolor}49.21 & \cellcolor{thirdcolor}48.21 & 40.47 & 40.14 & 47.61 & 23.00 & 14.53 & 6.33 & 25.21 \\
 & LOGIC & 62.28 & 61.48 & \cellcolor{firstcolor}\textbf{70.07} & \cellcolor{secondcolor}65.15 & 62.13 & 55.46 & 60.89 & \cellcolor{thirdcolor}64.53 & 43.40 & 24.87 & 17.07 & 46.94 \\
 & CONS & \cellcolor{thirdcolor}79.77 & 76.14 & \cellcolor{firstcolor}\textbf{85.33} & 79.17 & \cellcolor{secondcolor}79.87 & 71.20 & 76.91 & 72.20 & 52.60 & 77.67 & 68.57 & 69.58 \\
 & UI & \cellcolor{thirdcolor}83.26 & 79.26 & \cellcolor{firstcolor}\textbf{91.28} & \cellcolor{secondcolor}87.83 & 81.46 & 67.72 & 72.38 & 72.01 & 46.87 & 57.47 & 45.80 & 54.20 \\
 & QUAL & 78.10 & 73.07 & \cellcolor{secondcolor}87.87 & \cellcolor{firstcolor}\textbf{92.32} & \cellcolor{thirdcolor}82.46 & 52.00 & 66.80 & 65.26 & 35.47 & 43.40 & 27.83 & 40.00 \\
 & Overall & 68.75 & 66.40 & \cellcolor{firstcolor}\textbf{77.23} & \cellcolor{secondcolor}74.74 & \cellcolor{thirdcolor}70.83 & 57.37 & 63.42 & 64.32 & 40.27 & 43.59 & 33.12 & 47.19 \\ \midrule

\multirow{6}{*}{\centering\shortstack{real-app}} 
 & GOAL & 33.40 & 39.24 & \cellcolor{secondcolor}43.73 & \cellcolor{firstcolor}\textbf{43.83} & \cellcolor{secondcolor}43.73 & 41.34 & \cellcolor{thirdcolor}41.42 & 41.26 & 22.80 & 9.11 & 4.27 & 25.00 \\
 & LOGIC & 55.80 & \cellcolor{thirdcolor}59.32 & \cellcolor{secondcolor}61.72 & 59.06 & 57.20 & 55.47 & 50.25 & \cellcolor{firstcolor}\textbf{62.87} & 50.67 & 19.42 & 17.53 & 43.80 \\
 & CONS & 79.00 & \cellcolor{thirdcolor}80.48 & \cellcolor{firstcolor}\textbf{81.66} & \cellcolor{secondcolor}81.22 & 76.33 & 69.67 & 76.75 & 76.28 & 64.13 & 71.10 & 69.27 & 70.80 \\
 & UI & \cellcolor{thirdcolor}81.60 & 81.56 & \cellcolor{firstcolor}\textbf{88.40} & \cellcolor{secondcolor}87.35 & 78.00 & 64.93 & 71.75 & 76.54 & 50.73 & 46.77 & 42.33 & 58.80 \\
 & QUAL & 78.40 & 78.77 & \cellcolor{secondcolor}88.15 & \cellcolor{firstcolor}\textbf{92.40} & \cellcolor{thirdcolor}80.33 & 48.99 & 61.92 & 69.67 & 34.53 & 32.85 & 22.47 & 41.80 \\
 & Overall & 65.64 & \cellcolor{thirdcolor}67.87 & \cellcolor{secondcolor}72.73 & \cellcolor{firstcolor}\textbf{72.77} & 67.12 & 56.08 & 60.42 & 65.32 & 44.57 & 35.85 & 31.17 & 48.04 \\ \midrule

\multirow{6}{*}{\centering\shortstack{grounding}} 
 & GOAL & \cellcolor{firstcolor}\textbf{35.60} & \cellcolor{secondcolor}26.07 & 20.40 & 16.33 & \cellcolor{thirdcolor}21.55 & 17.73 & 15.85 & 17.40 & 8.47 & 7.89 & 6.73 & 5.82 \\
 & LOGIC & \cellcolor{firstcolor}\textbf{67.00} & \cellcolor{secondcolor}66.13 & 53.80 & 55.80 & 58.38 & 58.14 & 57.21 & 58.87 & 54.73 & 61.00 & \cellcolor{thirdcolor}63.87 & 62.07 \\
 & CONS & 78.20 & 85.00 & 75.82 & 59.33 & 75.35 & 68.04 & 84.69 & 79.80 & 67.07 & \cellcolor{thirdcolor}86.45 & \cellcolor{firstcolor}\textbf{91.53} & \cellcolor{secondcolor}89.03 \\
 & UI & 81.40 & 77.67 & 76.03 & 67.73 & 69.23 & 62.89 & 77.41 & 71.93 & 55.13 & \cellcolor{thirdcolor}85.45 & \cellcolor{firstcolor}\textbf{91.47} & \cellcolor{secondcolor}88.23 \\
 & QUAL & 80.40 & 76.53 & 80.74 & 75.93 & 75.76 & 63.30 & 82.18 & 77.53 & 55.27 & \cellcolor{secondcolor}87.17 & \cellcolor{firstcolor}\textbf{90.33} & \cellcolor{thirdcolor}84.01 \\
 & Overall & \cellcolor{secondcolor}68.52 & \cellcolor{thirdcolor}66.28 & 61.36 & 55.02 & 60.05 & 54.02 & 63.47 & 61.11 & 48.13 & 65.59 & \cellcolor{firstcolor}\textbf{68.79} & 65.83 \\ \bottomrule

\end{tabular}%
}
\end{table*}

\clearpage
\begin{table}[H]
\centering

\caption{Detailed GUI-Score results on GUI-GenBench with \textbf{Qwen3-VL-235B-A22B-Thinking}~\citep{bai2025qwen3vltechnicalreport} as the judge.}
\label{tab:model_evaluation_new}

\setlength{\tabcolsep}{1.5pt}
\renewcommand{\arraystretch}{0.9} % 行距压缩（非常关键）

\resizebox{\textwidth}{!}{%
\begin{tabular}{ll|cccccccccccc}
\toprule
\textbf{Task} & \textbf{Metric} & 
\begin{tabular}[c]{@{}c@{}}\textbf{Nano}\\\textbf{Banana Pro}\end{tabular} & 
\begin{tabular}[c]{@{}c@{}}\textbf{Nano}\\\textbf{Banana}\end{tabular} & 
\begin{tabular}[c]{@{}c@{}}\textbf{GPT-Image}\\\textbf{-1.5}\end{tabular} & 
\begin{tabular}[c]{@{}c@{}}\textbf{GPT-Image}\\\textbf{-1.0}\end{tabular} & 
\begin{tabular}[c]{@{}c@{}}\textbf{Seedream}\\\textbf{4.5}\end{tabular} & 
\begin{tabular}[c]{@{}c@{}}\textbf{Seedream}\\\textbf{4.0}\end{tabular} & 
\begin{tabular}[c]{@{}c@{}}\textbf{Wan 2.6}\end{tabular} & 
\begin{tabular}[c]{@{}c@{}}\textbf{Flux-2-Pro}\end{tabular} & 
\begin{tabular}[c]{@{}c@{}}\textbf{Bagel}\end{tabular} & 
\begin{tabular}[c]{@{}c@{}}\textbf{UniWorld}\end{tabular} & 
\begin{tabular}[c]{@{}c@{}}\textbf{Qwen}\\\textbf{-Image-Edit}\end{tabular} & 
\begin{tabular}[c]{@{}c@{}}\textbf{Longcat}\\\textbf{-Image}\end{tabular} \\ \midrule

\multirow{6}{*}{\centering\shortstack{single-step}} 
 & GOAL & \cellcolor{firstcolor}\textbf{93.64} & 70.54 & \cellcolor{secondcolor}84.70 & \cellcolor{thirdcolor}75.53 & 67.24 & 65.85 & 67.65 & 66.84 & 27.14 & 38.72 & 33.23 & 35.05 \\
 & LOGIC & \cellcolor{firstcolor}\textbf{92.45} & \cellcolor{thirdcolor}72.86 & \cellcolor{secondcolor}81.27 & 71.20 & 66.02 & 64.41 & 69.13 & 68.27 & 30.44 & 38.32 & 35.32 & 34.78 \\
 & CONS & \cellcolor{firstcolor}\textbf{83.28} & 66.40 & 64.60 & 55.43 & 59.66 & 57.68 & \cellcolor{secondcolor}70.31 & \cellcolor{thirdcolor}69.97 & 37.58 & 36.50 & 44.07 & 34.88 \\
 & UI & \cellcolor{firstcolor}\textbf{94.25} & 65.62 & \cellcolor{secondcolor}83.98 & \cellcolor{thirdcolor}70.14 & 62.41 & 61.16 & 65.02 & 63.57 & 31.85 & 40.17 & 44.78 & 35.59 \\
 & QUAL & \cellcolor{firstcolor}\textbf{98.04} & 80.51 & \cellcolor{secondcolor}94.85 & \cellcolor{thirdcolor}86.80 & 72.59 & 76.03 & 77.37 & 80.58 & 41.21 & 55.19 & 60.30 & 48.48 \\
 & Overall & \cellcolor{firstcolor}\textbf{92.33} & \cellcolor{thirdcolor}71.19 & \cellcolor{secondcolor}81.88 & 71.82 & 65.58 & 65.03 & 69.90 & 69.85 & 33.64 & 41.78 & 43.54 & 37.76 \\ \midrule

\multirow{6}{*}{\centering\shortstack{multi-step}} 
 & GOAL & \cellcolor{thirdcolor}92.35 & 60.97 & \cellcolor{secondcolor}93.61 & \cellcolor{firstcolor}\textbf{96.22} & 89.39 & 83.45 & 72.70 & 83.80 & 4.62 & 6.44 & 4.22 & 3.70 \\
 & LOGIC & \cellcolor{secondcolor}88.53 & 60.44 & \cellcolor{thirdcolor}84.43 & \cellcolor{firstcolor}\textbf{89.46} & 77.59 & 75.09 & 65.92 & 76.15 & 17.65 & 23.95 & 19.70 & 16.69 \\
 & CONS & \cellcolor{firstcolor}\textbf{84.04} & 71.32 & 75.50 & \cellcolor{secondcolor}79.22 & 73.23 & 71.26 & \cellcolor{thirdcolor}76.32 & 72.29 & 19.69 & 42.06 & 57.16 & 30.29 \\
 & UI & \cellcolor{firstcolor}\textbf{95.96} & 60.77 & \cellcolor{thirdcolor}90.68 & \cellcolor{secondcolor}92.31 & 75.00 & 73.23 & 74.48 & 64.74 & 16.12 & 33.32 & 51.10 & 21.16 \\
 & QUAL & \cellcolor{secondcolor}97.65 & 71.89 & \cellcolor{thirdcolor}96.34 & \cellcolor{firstcolor}\textbf{98.20} & 86.87 & 80.17 & 87.13 & 77.29 & 17.65 & 48.06 & 53.59 & 21.16 \\
 & Overall & \cellcolor{firstcolor}\textbf{91.71} & 65.08 & \cellcolor{thirdcolor}88.11 & \cellcolor{secondcolor}91.08 & 80.42 & 76.64 & 75.31 & 74.85 & 15.15 & 30.77 & 37.15 & 18.60 \\ \midrule

\multirow{6}{*}{\centering\shortstack{fictional-app}} 
 & GOAL & 28.13 & 34.48 & \cellcolor{firstcolor}\textbf{43.33} & \cellcolor{thirdcolor}34.68 & \cellcolor{secondcolor}39.19 & 34.27 & 27.50 & 34.13 & 15.69 & 9.73 & 8.07 & 18.40 \\
 & LOGIC & 41.93 & \cellcolor{secondcolor}49.90 & \cellcolor{firstcolor}\textbf{58.33} & 46.87 & \cellcolor{thirdcolor}48.26 & 45.67 & 45.28 & 48.70 & 30.24 & 19.60 & 13.93 & 31.39 \\
 & CONS & \cellcolor{thirdcolor}49.47 & \cellcolor{firstcolor}\textbf{58.05} & \cellcolor{secondcolor}52.87 & 43.30 & 49.26 & 43.73 & 49.51 & 43.33 & 25.86 & 53.07 & 20.53 & 36.04 \\
 & UI & 53.47 & \cellcolor{thirdcolor}55.69 & \cellcolor{firstcolor}\textbf{70.20} & \cellcolor{secondcolor}63.91 & 55.64 & 45.27 & 47.64 & 40.67 & 22.36 & 34.00 & 19.47 & 29.93 \\
 & QUAL & 76.53 & 72.73 & \cellcolor{firstcolor}\textbf{97.00} & \cellcolor{secondcolor}94.75 & \cellcolor{thirdcolor}82.21 & 60.47 & 73.12 & 63.00 & 20.88 & 34.00 & 24.47 & 29.03 \\
 & Overall & 49.91 & 54.17 & \cellcolor{firstcolor}\textbf{64.35} & \cellcolor{secondcolor}56.70 & \cellcolor{thirdcolor}54.91 & 45.88 & 48.61 & 45.97 & 23.01 & 30.08 & 17.29 & 28.96 \\ \midrule

\multirow{6}{*}{\centering\shortstack{real-app}} 
 & GOAL & 30.47 & 30.47 & 34.73 & \cellcolor{secondcolor}37.33 & \cellcolor{thirdcolor}36.67 & \cellcolor{firstcolor}\textbf{39.00} & 27.95 & 35.08 & 17.58 & 8.40 & 4.33 & 21.73 \\
 & LOGIC & 43.47 & 47.32 & \cellcolor{firstcolor}\textbf{50.73} & \cellcolor{secondcolor}49.40 & 45.87 & 46.27 & 47.03 & \cellcolor{thirdcolor}48.56 & 33.60 & 14.95 & 12.80 & 31.33 \\
 & CONS & \cellcolor{secondcolor}58.33 & \cellcolor{firstcolor}\textbf{62.21} & \cellcolor{thirdcolor}53.13 & 49.93 & 48.73 & 42.80 & 51.30 & 47.81 & 34.75 & 37.75 & 17.60 & 43.96 \\
 & UI & \cellcolor{thirdcolor}59.80 & 59.53 & \cellcolor{firstcolor}\textbf{73.07} & \cellcolor{secondcolor}65.47 & 50.47 & 41.53 & 50.88 & 48.42 & 26.06 & 22.39 & 15.13 & 34.93 \\
 & QUAL & 75.60 & 78.52 & \cellcolor{firstcolor}\textbf{97.93} & \cellcolor{secondcolor}93.67 & \cellcolor{thirdcolor}80.13 & 56.00 & 67.03 & 66.94 & 20.88 & 22.73 & 16.60 & 34.53 \\
 & Overall & 53.53 & \cellcolor{thirdcolor}55.61 & \cellcolor{firstcolor}\textbf{61.92} & \cellcolor{secondcolor}59.16 & 52.37 & 45.12 & 48.84 & 49.36 & 26.57 & 21.24 & 13.29 & 33.30 \\ \midrule

\multirow{6}{*}{\centering\shortstack{grounding}} 
 & GOAL & \cellcolor{firstcolor}24.38 & \cellcolor{secondcolor}20.89 & 15.68 & 18.72 & \cellcolor{thirdcolor}19.31 & 14.41 & 16.05 & 16.58 & 12.44 & 12.50 & 6.69 & 9.08 \\
 & LOGIC & \cellcolor{thirdcolor}36.50 & \cellcolor{secondcolor}36.56 & 32.26 & 30.54 & \cellcolor{firstcolor}\textbf{37.66} & 32.21 & 34.45 & 31.81 & 22.34 & 22.85 & 15.00 & 18.16 \\
 & CONS & \cellcolor{thirdcolor}46.20 & 42.89 & 33.77 & 25.10 & 40.96 & 38.76 & \cellcolor{firstcolor}\textbf{48.21} & \cellcolor{secondcolor}46.44 & 35.79 & 40.83 & 42.04 & 42.13 \\
 & UI & \cellcolor{firstcolor}\textbf{69.49} & 58.11 & 56.85 & 44.30 & 56.22 & 52.55 & \cellcolor{thirdcolor}67.22 & 56.98 & 40.07 & 59.03 & \cellcolor{secondcolor}68.66 & 61.35 \\
 & QUAL & \cellcolor{firstcolor}\textbf{89.12} & 79.38 & 81.10 & 70.40 & 76.29 & 72.76 & \cellcolor{secondcolor}86.08 & 73.56 & 53.11 & 81.39 & \cellcolor{thirdcolor}84.23 & 80.21 \\
 & Overall & \cellcolor{firstcolor}\textbf{53.14} & \cellcolor{thirdcolor}47.57 & 43.93 & 37.81 & 46.09 & 42.14 & \cellcolor{secondcolor}50.40 & 45.07 & 32.75 & 43.32 & 43.32 & 42.19 \\ \bottomrule

\end{tabular}%
}
\end{table}

\section{Task-Specific Scoring Rubrics}
\label{sec:rubrics}

\begin{figure*}[t]
    \centering
    \includegraphics[width=0.96\textwidth]{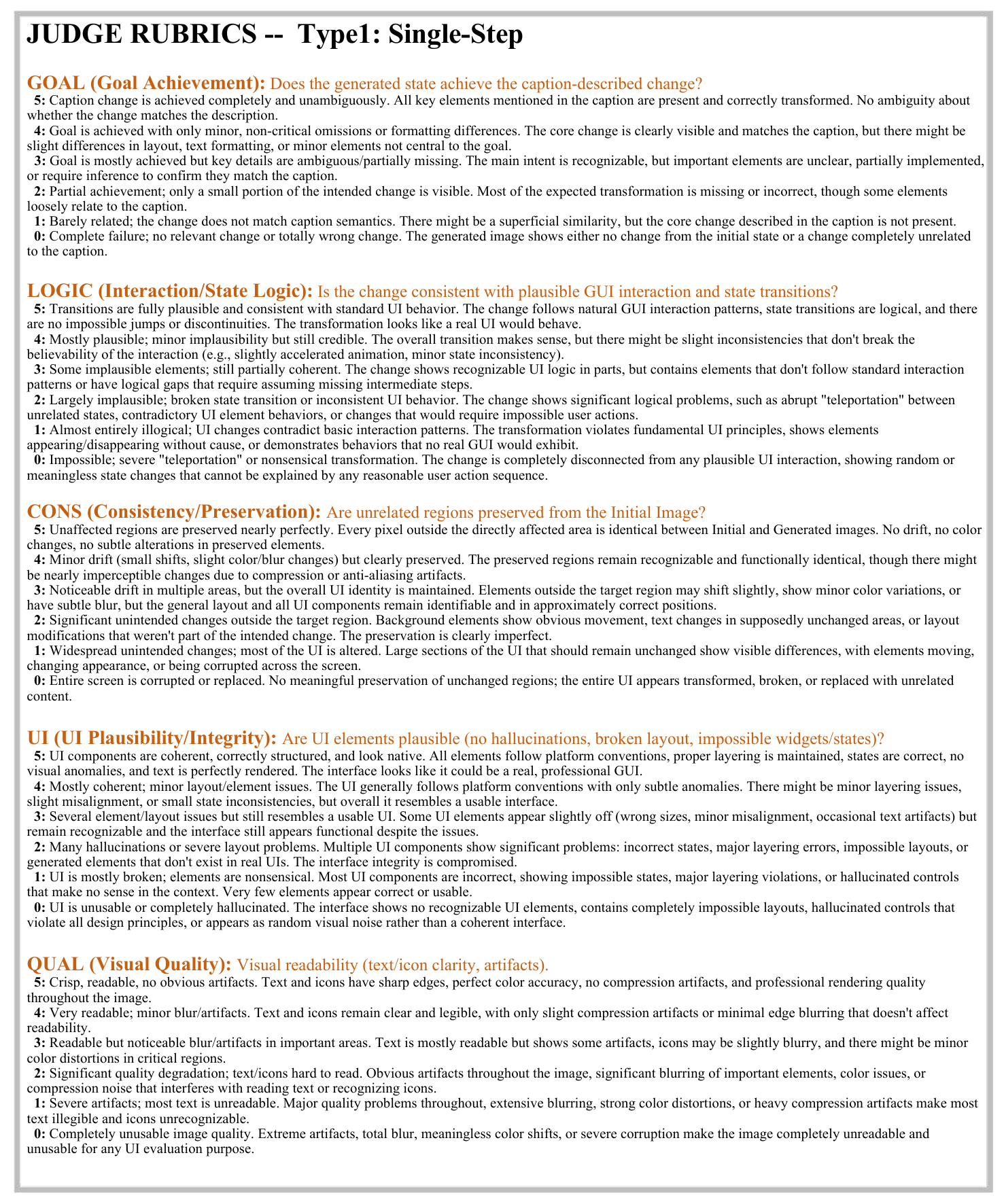}
    \caption{GUI-Score rubric for Single-Step Visual Transition.}
    \label{fig:judge_ss}
\end{figure*}

\begin{figure*}[t]
    \centering
    \includegraphics[width=0.96\textwidth]{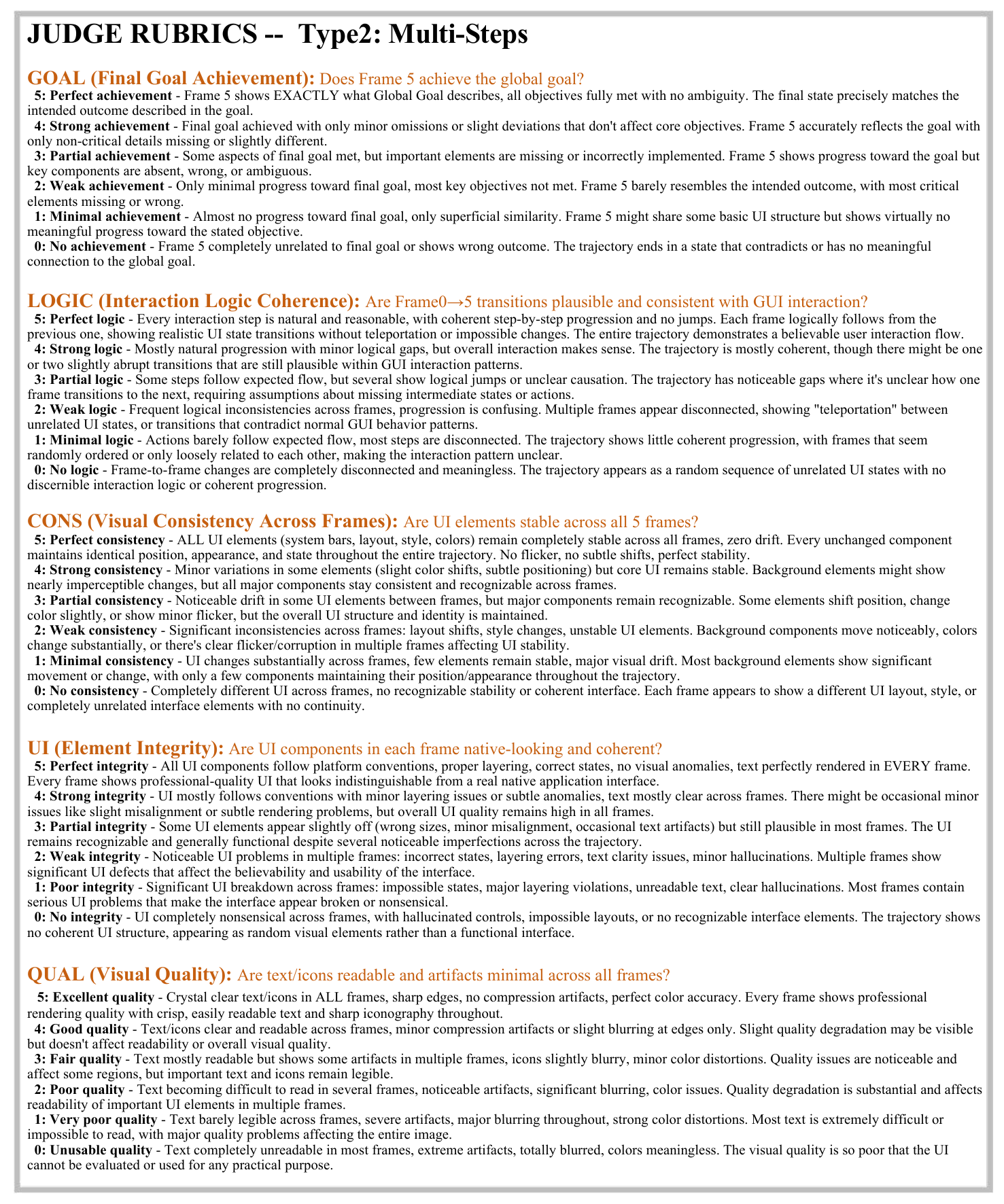}
    \caption{GUI-Score rubric for Multi-Step Planning.}
    \label{fig:judge_ms}
\end{figure*}

\begin{figure*}[t]
    \centering
    \includegraphics[width=0.96\textwidth]{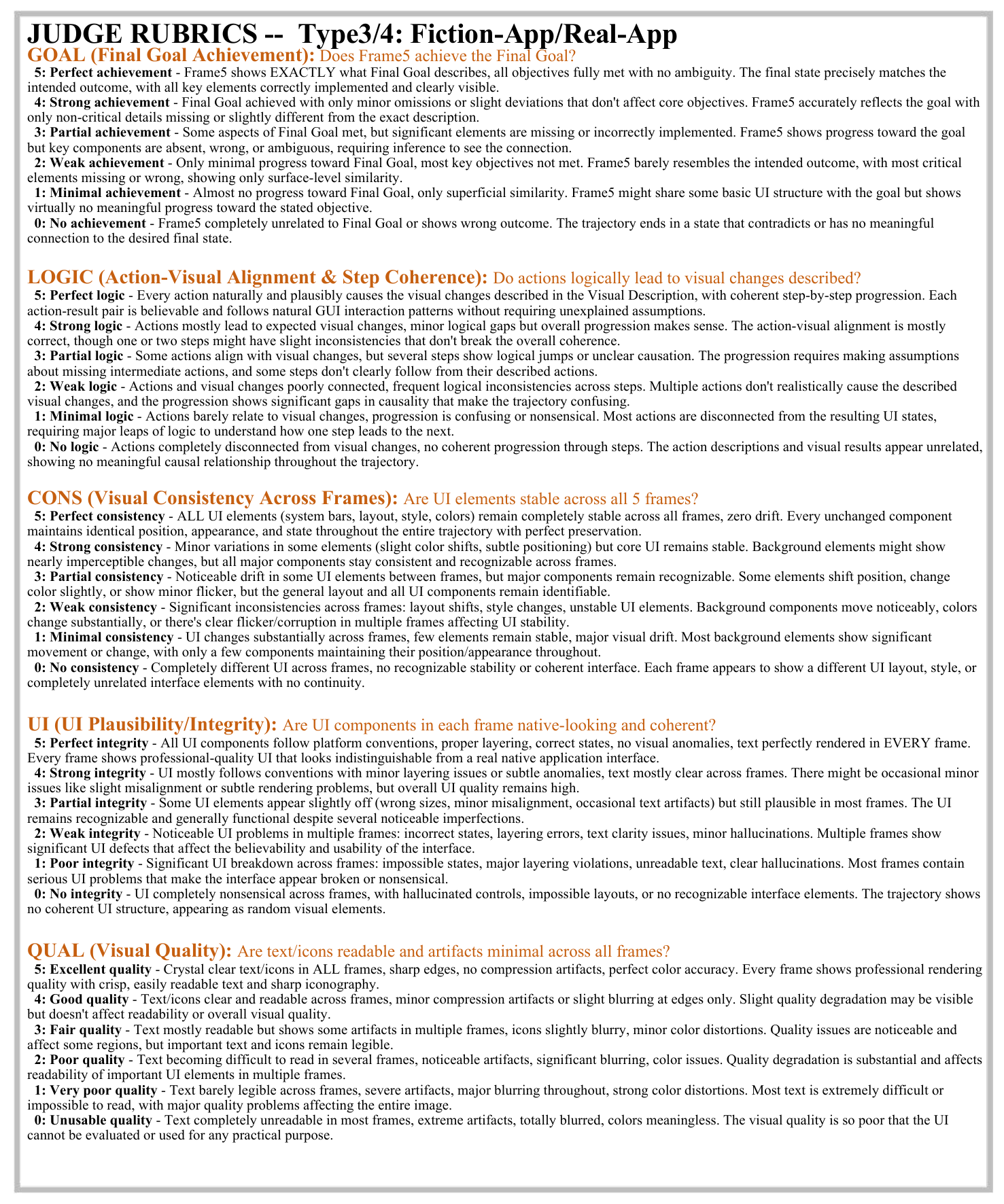}
    \caption{GUI-Score rubrics for Zero-shot Virtual GUI and Rare Trajectory Synthesis.}
    \label{fig:judge_tt}
\end{figure*}

\begin{figure*}[t]
    \centering
    \includegraphics[width=0.96\textwidth]{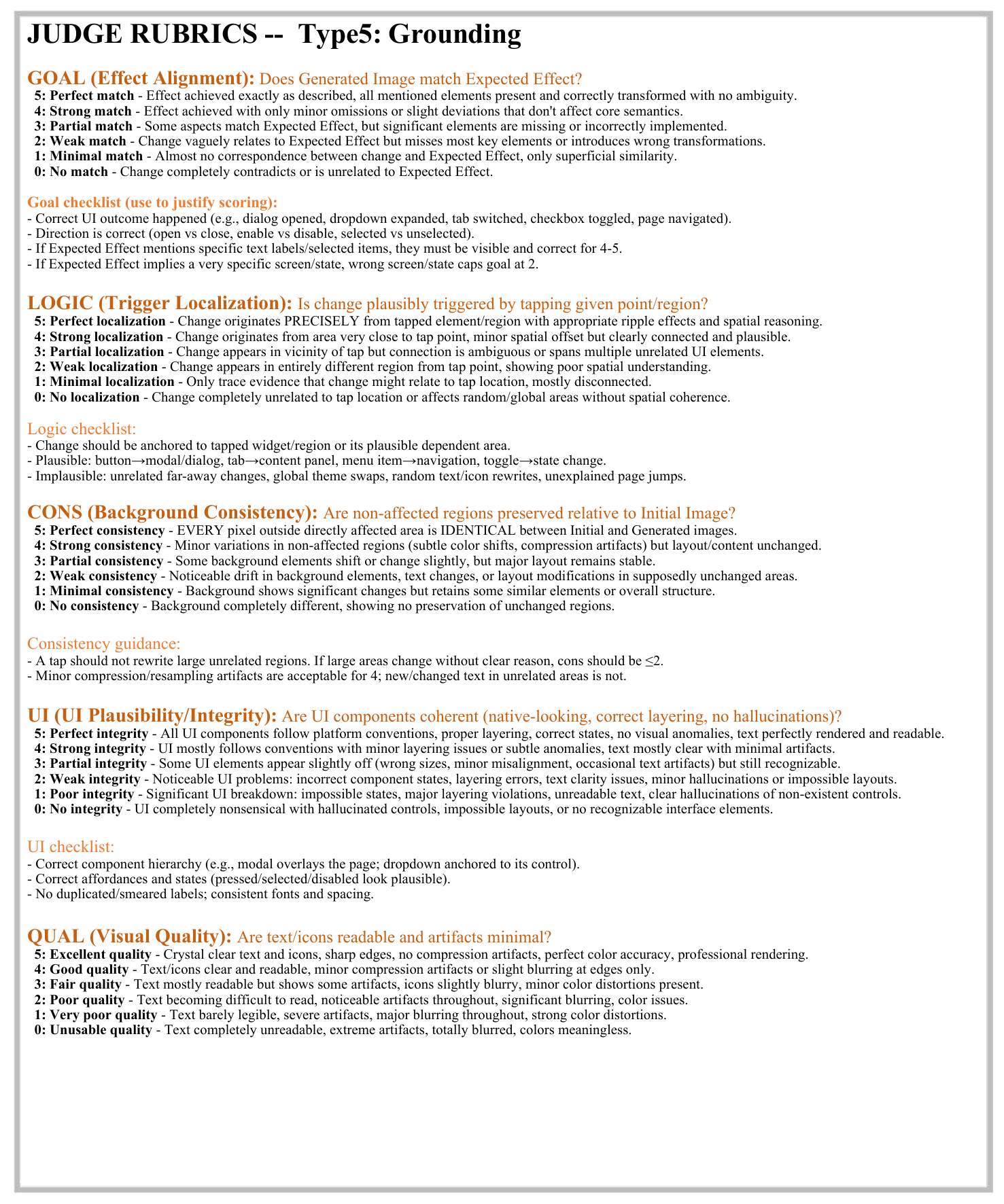}
    \caption{GUI-Score rubric for Grounding-based Generation.}
    \label{fig:app_gd}
\end{figure*}

Figures~\ref{fig:judge_ss}, \ref{fig:judge_ms}, \ref{fig:judge_tt}, and \ref{fig:app_gd} reproduce the rubric cards used by the VLM judges. The five task suites share the GOAL, LOGIC, CONS, UI, and QUAL dimensions, while the descriptions specify how each dimension applies to the task. Performance tiers range from ``Incomplete'' to ``Exceptional'':

\begin{itemize} 
    \item Figure~\ref{fig:judge_ss}: single-step transitions, including whether the generated screen realizes the requested action.
    \item Figure~\ref{fig:judge_ms}: multi-step planning, including coherent progression and consistency across five-step trajectories.
    \item Figure~\ref{fig:judge_tt}: fictional-app generation and real-app trajectories, including structural plausibility and consistency in less familiar settings.
    \item Figure~\ref{fig:app_gd}: grounding, including whether the resulting state corresponds to the coordinate-specified interaction.
\end{itemize}

The cards specify the qualitative basis of the ordinal ratings. They should be read alongside the generation protocol and the judge-specific results, rather than as guarantees of error-free evaluation.

% \begin{figure}
%     \centering
%     \includegraphics[width=1\linewidth]{./figures/distribution.png}
%     \caption{Enter Caption}
%     \label{fig:placeholder}
% \end{figure}

\end{document}